%% file: swe_serve_arxiv.tex
\documentclass{article} % For LaTeX2e
\usepackage{iclr2027_conference,times}

\input{math_commands.tex}

\usepackage{microtype}
\usepackage{booktabs}
\usepackage{colortbl}
\usepackage{array}
\usepackage{tabularx}
\usepackage{amssymb}
\usepackage{graphicx}
\usepackage{xcolor}
\usepackage{float}
\usepackage{placeins}
\usepackage{longtable}
\usepackage{hyperref}
\usepackage{url}

\definecolor{sweservebluelight}{HTML}{DCE6F1}

\graphicspath{{figures/}}
\hypersetup{
  pdftitle={SWE-Serve: Benchmarking Agentic Engineering for Production Inference Serving},
  pdfauthor={},
  pdfsubject={Benchmarking agentic engineering for production inference serving},
  pdfkeywords={coding agents, software engineering benchmarks, inference serving, GPU systems}
}

\newcommand{\FtwoP}{\ensuremath{\mathrm{F2P}}}
\newcommand{\PtwoP}{\ensuremath{\mathrm{P2P}}}
\newcommand{\featurecheck}{\ensuremath{\checkmark}}

\input{generated/benchmark_facts}


\title{SWE-Serve: Benchmarking Agentic Engineering for Production Inference Serving}

\author{%
  \parbox[t]{\dimexpr\textwidth-2\tabcolsep\relax}{%
    \centering
    Jennifer Williams\textsuperscript{1}\thanks{Correspondence:
      \href{mailto:jewilliams@nvidia.com}{\nolinkurl{jewilliams@nvidia.com}}}\quad
    Dave Farris\textsuperscript{1}\quad
    Jeff Farris\textsuperscript{1}\quad
    Jiantao Jiao\textsuperscript{1,2}\\[2pt]
    \normalfont
    \textsuperscript{1}NVIDIA\qquad
    \textsuperscript{2}University of California, Berkeley
  }%
}
\hypersetup{pdfauthor={Jennifer Williams, Dave Farris, Jeff Farris, Jiantao Jiao}}

\iclrfinalcopy
\usepackage{etoolbox}
\makeatletter
\patchcmd{\@maketitle}
  {\lhead{Published as a conference paper at ICLR 2027}}
  {}
  {}
  {\PackageError{sweserve-preprint}{Could not remove conference publication header}
    {Check the title macro in the conference style before exporting.}}
\patchcmd{\@maketitle}
  {\vskip 0.3in minus 0.1in}
  {\vskip 0.12in}
  {}
  {\PackageError{sweserve-preprint}{Could not adjust title-block spacing}
    {Check the title macro in the conference style before exporting.}}
\makeatother
\renewcommand{\headrulewidth}{0pt}
\hypersetup{pdfsubject={Research preprint}}

\begin{document}

% The official instructional sample body is replaced by the manuscript below,
% as directed by the upstream shell. It remains verbatim in the vendored copy.
% BEGIN SWE-SERVE PAPER CONTENT

\maketitle

\begin{abstract}
\input{iclr2027_sections/00_abstract}
\end{abstract}

% Keep later section floats from jumping ahead of the title and abstract.
\FloatBarrier

\input{iclr2027_sections/01_introduction}
\input{iclr2027_sections/02_positioning}
\input{iclr2027_sections/03_benchmark}
\input{iclr2027_sections/04_experimental_setup}
\input{iclr2027_sections/05_results}
\input{iclr2027_sections/06_production_contract}
\input{iclr2027_sections/07_limitations_conclusion}

% Hold all main-paper floats above the page-limit marker. The build script reads this marker.
\FloatBarrier
\typeout{ICLR-MAIN-TEXT-END-PAGE=\thepage}

% Begin material outside the main-text page limit on a new page.
\clearpage
\input{iclr2027_sections/90_statements}

\appendix
\input{iclr2027_sections/99_appendix}

% END SWE-SERVE PAPER CONTENT
\end{document}

%% file: math_commands.tex
\usepackage{amsmath,amsfonts,bm}

\def\eqref#1{equation~\ref{#1}}
\def\1{\bm{1}}

\DeclareMathAlphabet{\mathsfit}{\encodingdefault}{\sfdefault}{m}{sl}
\SetMathAlphabet{\mathsfit}{bold}{\encodingdefault}{\sfdefault}{bx}{n}

%% file: generated/benchmark_facts.tex
\newcommand{\BenchmarkTaskCount}{53}
\newcommand{\BenchmarkCpuTaskCount}{12}
\newcommand{\BenchmarkHOneHundredTaskCount}{41}

\newcommand{\BenchmarkPerformanceTaskCount}{3}
\newcommand{\BenchmarkSingleChangeTaskCount}{37}
\newcommand{\BenchmarkMultiChangeTaskCount}{16}

\newcommand{\BenchmarkUniqueSourceChangeCount}{83}

%% file: iclr2027_sections/00_abstract.tex
We introduce SWE-Serve, a benchmark for evaluating agents on production inference engineering
tasks. Implementing an inference feature can require coordinating multiple changes across the
serving stack, including model support, runtime execution, and public APIs. Existing benchmarks
provide limited coverage of production inference engineering: repository-level software engineering
benchmarks do not target inference, while general terminal-agent benchmarks include only
a few inference tasks. Dedicated inference benchmarks, meanwhile, focus primarily on isolated
kernel generation or performance optimization rather than repository-scale production feature
implementation.
SWE-Serve provides 53 repository-grounded tasks derived from recent production changes to SGLang,
spanning six inference engineering families. Each task executes on either CPU or a single GPU (H100)
and is evaluated with hidden functional and regression tests, including, where applicable,
end-to-end (E2E) serving tests and calibrated performance gates. Executable no-op and oracle controls, adversarial
verifier review, and closed-book execution support task validity and evaluation integrity. Across 11
models and 31 model--effort configurations, the best-performing configuration achieves 75\% mean
pass@1. SWE-Serve exposes a substantial gap between completing tasks locally and achieving
production correctness. On 19 tasks with end-to-end
coverage, model-serving E2E tests reject roughly one-third of patches that pass every other test
(45.9\% under the verifier versus 69.4\% with E2E tests excluded from scoring), with pass rate
increasing for each model's best-performing configuration. By making the production correctness gap
directly measurable, SWE-Serve enables the field to track whether future agents move beyond
completing tasks locally to achieving production correctness.

%% file: iclr2027_sections/01_introduction.tex
\section{Introduction}

Repository-scale software engineering has become a common setting for coding-agent
evaluation~\citep{jimenez2024swebench,openai2024swebenchverified,deng2025swebenchpro,
huang2026deepswe}. Recent benchmarks have begun to extend this setting toward feature-scale and
longer-horizon work~\citep{zhou2026featurebench,feng2026longcli,luo2026rpg,
yang2026programbench,thai2025sweevo}. However, repository-scale
evaluation does not necessarily test whether a locally working implementation remains correct
across the production path. We introduce SWE-Serve, which evaluates whether agents achieve
production correctness for inference engineering tasks.

In production inference systems, a change can work in isolation yet fail when exercised through the
complete system. On the Gemma 4 mixture-of-experts (MoE) core-serving task, for example, 48.5\% of
the agent-created patches evaluated for the primary leaderboard (16 of 33) passed every test other than the
end-to-end (E2E) tests but failed to serve the specified model correctly through the live server's
public interfaces (Appendix
Table~\ref{tab:strict-e2e-task-decomposition}). Other SWE-Serve tasks test whether patches remain
correct when requests share persistent state or execute concurrently. These failures affect users
despite apparent local correctness, motivating us to define and measure \emph{production correctness}:
satisfying a task's behavioral requirements within the inference system, including preserving existing
functionality and meeting any task-specific performance constraints.

SWE-Serve is a benchmark for production inference engineering built from changes merged since
December 2025 into SGLang, an open-source inference-serving system~\citep{zheng2024sglang}. There are 53
production-grounded tasks spanning six families of inference engineering, from model support and
decoding to distributed execution and serving APIs. Tasks focus on concrete engineering objectives,
derived either from individual pull requests (PRs) or from combinations of related PRs when the work
spans PR boundaries.
Each task consists of a containerized environment with the repository checked out at the task's
designated base commit, a task instruction, hidden executable tests, and an oracle solution that
establishes task feasibility. Each task executes in either a CPU-only environment or on a single
H100 GPU. We qualify each task by checking that its hidden tests reject the unchanged repository
(the no-op control) and accept the oracle solution, followed by agent-assisted adversarial probing
and human adjudication. Additionally, evaluation integrity is
protected with closed-book execution and trajectory auditing.

SWE-Serve provides an executable, production-grounded evaluation of whether agents carry locally
working implementations through a complete production path. We release the benchmark, execution
environments, and verifiers to support reproducible evaluation. Our contributions are:

\begin{itemize}
  \item \textbf{A production inference engineering benchmark:} 53 production-grounded tasks across
  six engineering families, with CPU/H100 execution and hidden functional and regression tests.
  \item \textbf{A validity and integrity methodology} combining production-grounded construction,
  executable no-op and oracle controls, agent-assisted adversarial probing, human
  adjudication, closed-book execution, and trajectory auditing.
  \item \textbf{Evidence of a production correctness gap.} Removing model-serving end-to-end (E2E)
  tests from scoring while leaving the agent-created patches and every other test unchanged raises
  pass rate by 23.4 percentage points, with an increase in all 11 top-per-model configurations.
  Among the 12 tasks that support a matched comparison, removing E2E tests produces 2.0$\times$ as
  many newly passing patches as removing the same number and type of non-E2E tests.
\end{itemize}

%% file: iclr2027_sections/02_positioning.tex
\section{Related Work}

Existing benchmarks cover only narrow slices of production inference engineering. SWE-Serve builds
upon related work spanning repository-level agentic benchmarks, inference-performance benchmarks,
task construction across related repository changes, and closed-book evaluation
(Table~\ref{tab:benchmark-landscape}).

\paragraph{Agentic benchmarks.}
Repository-level software engineering benchmarks, such as SWE-bench Verified, SWE-bench Pro, and
DeepSWE, have established executable evaluation of agents on changes to existing software
repositories~\citep{jimenez2024swebench,openai2024swebenchverified,deng2025swebenchpro,
huang2026deepswe}. However, they do not focus on inference engineering or GPU-dependent execution.
Terminal-Bench 3 broadens agentic evaluation to interactive systems tasks and introduces a few
inference tasks (3 of 74 tasks; 4.1\%), but they remain a minor component of the
benchmark~\citep{merrill2026terminalbench,marten2026frontierbench}. Together, these benchmarks
provide limited coverage of the breadth of inference engineering.

\paragraph{Inference-performance benchmarks.}
Existing inference-performance benchmarks target several distinct layers of the inference stack.
KernelBench and SOL-ExecBench evaluate GPU-kernel generation and optimization, while InferenceBench
targets deployed-system inference optimization~\citep{ouyang2025kernelbench,lin2026solexecbench,
yeon2026inferencebench}. ISO-Bench is the only benchmark in this group that evaluates inference
optimization at the repository level; its tasks are drawn exclusively from performance-optimization
changes in vLLM and SGLang~\citep{nangia2026isobench}. However, inference engineering encompasses a
broader range of tasks than performance alone.

\paragraph{Construction across PRs.}
Pull requests reflect repository-specific development practices rather than a standardized unit of
functionality: a coherent engineering objective may be contained in one PR or distributed across
several. FeatureBench and SWE-EVO demonstrate task constructions that group related repository
changes rather than requiring every task to follow a single-PR
boundary~\citep{zhou2026featurebench,thai2025sweevo}. SWE-Serve brings this flexibility to production
inference engineering, scoping one or more related upstream changes into bounded, executable tasks.
This permits work that spans PR boundaries to be evaluated together.

\paragraph{Closed-book evaluation.}
Closed-book execution, where agents cannot retrieve upstream solutions from the public web,
remains uncommon among agentic benchmarks. DeepSWE and SWE-EVO are notable
exceptions~\citep{huang2026deepswe,datacurve2026deepswerepo,thai2025sweevo}. SWE-Serve similarly
evaluates agents closed book and further audits every evaluation trajectory for prohibited access.

\begin{table}[H]
  \caption{Comparison of SWE-Serve with related benchmarks. Checkmarks indicate documented coverage.
  Labels denote: ``Beyond performance'': inference-engineering tasks beyond optimization;
  ``Closed-book'': execution in which agents cannot retrieve upstream solutions; and ``Multi-PR'':
  tasks constructed from multiple PRs. Only 3 of 74 Terminal-Bench 3 tasks concern inference
  engineering.}
  \label{tab:benchmark-landscape}
  \begin{center}
  \normalsize
  \setlength{\tabcolsep}{2pt}
  \renewcommand{\arraystretch}{1.03}
  \begin{tabular}{@{}l*{8}{c}@{}}
    \toprule
    & \multicolumn{3}{c}{Inference coverage}
      & \multicolumn{2}{c}{Task construction}
      & \multicolumn{3}{c}{Execution setting} \\
    \cmidrule(lr){2-4}\cmidrule(lr){5-6}\cmidrule(l){7-9}
    Benchmark & \shortstack{\strut Inference\\\strut focused} & \shortstack{\strut Perf.\\\strut metrics}
      & \shortstack{\strut Beyond\\\strut perf.} & \shortstack{\strut Repo-\\\strut grounded}
      & \shortstack{\strut Multi-\\\strut PR} & \shortstack{\strut GPU\\\strut task}
      & \shortstack{\strut CPU-only\\\strut task} & \shortstack{\strut Closed-\\\strut book} \\
    \midrule
\input{generated/benchmark_landscape_rows.tex} 
    \bottomrule
  \end{tabular}
  \end{center}
\end{table}

\FloatBarrier

%% file: generated/benchmark_landscape_rows.tex
% Generated by scripts/paper/generate_benchmark_landscape_table.py; do not edit by hand.
\multicolumn{9}{@{}l}{\textit{General SWE}} \\[-1pt]
\quad SWE-bench Verified &  &  &  & \featurecheck &  &  & \featurecheck &  \\
\quad SWE-bench Pro &  &  &  & \featurecheck &  &  & \featurecheck &  \\
\quad DeepSWE &  &  &  & \featurecheck &  &  & \featurecheck & \featurecheck \\
\addlinespace[1pt]
\multicolumn{9}{@{}l}{\textit{Terminal agents}} \\[-1pt]
\quad Terminal-Bench 3 &  & \featurecheck & \featurecheck & \featurecheck &  & \featurecheck & \featurecheck &  \\
\addlinespace[1pt]
\multicolumn{9}{@{}l}{\textit{Inference performance}} \\[-1pt]
\quad KernelBench & \featurecheck & \featurecheck &  &  &  & \featurecheck &  &  \\
\quad SOL-ExecBench & \featurecheck & \featurecheck &  &  &  & \featurecheck &  &  \\
\quad ISO-Bench & \featurecheck & \featurecheck &  & \featurecheck &  & \featurecheck &  &  \\
\quad InferenceBench & \featurecheck & \featurecheck &  &  &  & \featurecheck &  &  \\
\addlinespace[1pt]
\multicolumn{9}{@{}l}{\textit{Production inference}} \\[-1pt]
\rowcolor{sweservebluelight}
\quad \textbf{SWE-Serve} & \featurecheck & \featurecheck & \featurecheck & \featurecheck & \featurecheck & \featurecheck & \featurecheck & \featurecheck \\

%% file: iclr2027_sections/03_benchmark.tex
\section{The SWE-Serve Benchmark}
\label{sec:benchmark}

\begin{figure}[t]
  \centering
  \includegraphics[width=\linewidth]{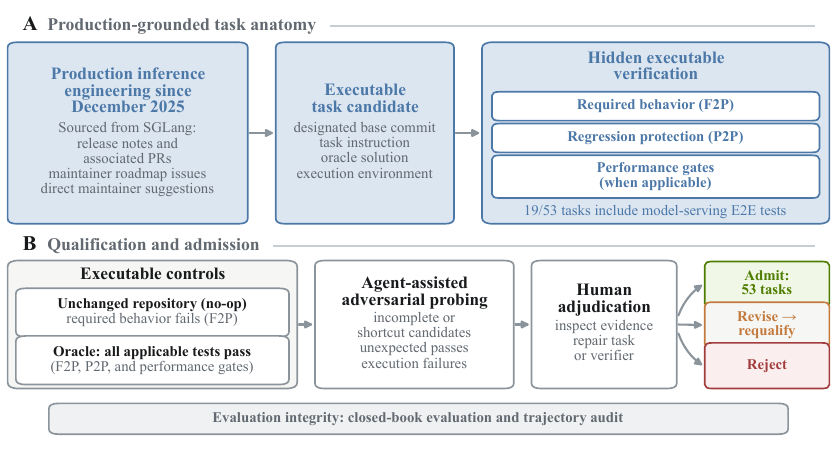}
  \caption{\textbf{SWE-Serve task construction and qualification.} Production grounding,
  executable verification, and qualification jointly support task validity. \textbf{(A)} Recent
  production inference engineering work from SGLang is mapped to executable task candidates with
  hidden behavioral verification. \textbf{(B)} Task admission is informed by no-op and oracle controls,
  agent-assisted adversarial probing, and human adjudication.}
  \label{fig:benchmark-anatomy}
\end{figure}

SWE-Serve turns recent engineering work on production inference systems into 53 repository-level
software engineering tasks with executable behavioral verification. We build the benchmark from
merged changes to SGLang~\citep{zheng2024sglang}, an open-source inference-serving system. As
Figure~\ref{fig:benchmark-anatomy} illustrates, each task gives an agent a task instruction and the
SGLang codebase at the task's designated base commit. The agent creates a patch; hidden
executable tests then evaluate the patch for required behavior, regression protection, and applicable
end-to-end or performance requirements. This section defines the tasks, characterizes the resulting
benchmark, and presents the construction and qualification evidence supporting its validity.

\subsection{Task Formulation and Scoring}
\label{sec:task-formulation-scoring}

Each SWE-Serve task includes an instruction, a sandboxed execution environment specification, an
oracle solution, and a hidden executable verifier. We package and execute these tasks using Harbor,
an open-source agent-evaluation framework from the creators of
Terminal-Bench~\citep{harbor2026,merrill2026terminalbench}. During execution, the agent may inspect,
edit, build, and test the repository, including its existing tests. It does not receive the task's oracle
solution, source-change identifiers, or the hidden verifier.

After the agent's work is completed, the hidden verifier evaluates the resulting agent-created patch.
The agent-created patch is not directly compared with the oracle patch, which is used only to qualify
that the task contract is achievable. Instead, following a standard established in SWE-bench, the
verifier assesses behavior through executable tests: fail-to-pass (F2P) tests require behavior absent
at the task's designated base commit, while pass-to-pass (P2P) tests protect relevant behavior already
present there~\citep{jimenez2024swebench}. When required by the task, the verifier additionally
evaluates model-serving end-to-end behavior or enforces calibrated performance gates. The task's
complete set of tests defines its \emph{production correctness} requirements, and an agent passes only if every test
passes. Any implementation that passes those tests is accepted. An implementation that fails any
F2P or P2P test therefore does not satisfy the task's production correctness requirements.

\subsection{Composition and Scope}
\label{sec:composition-scope}

Table~\ref{tab:benchmark-scale} summarizes engineering breadth and task scale. To give a compact view
of the engineering work represented, we organize the tasks into six mutually exclusive engineering
families (\hyperref[tab:benchmark-scale-a]{Table~\ref*{tab:benchmark-scale}A}).
The tasks also vary in construction scope:
\BenchmarkSingleChangeTaskCount{} derive from one upstream change,
while \BenchmarkMultiChangeTaskCount{} integrate two to six related changes. In total, the benchmark
draws on \BenchmarkUniqueSourceChangeCount{} unique merged upstream pull requests. Execution
spans \BenchmarkCpuTaskCount{} CPU and \BenchmarkHOneHundredTaskCount{} GPU (H100) tasks, with strict
model-serving E2E tests in 19 tasks and calibrated performance gates in
\BenchmarkPerformanceTaskCount{}. Together, the benchmark spans localized correctness fixes and
integrations that cross API, scheduling, model-execution, and runtime-state boundaries.

\begin{table}[!htbp]
  \caption{SWE-Serve benchmark composition and scale.}
  \label{tab:benchmark-scale}
  \begin{center}
  \normalsize
  \begin{minipage}[t]{0.50\linewidth}
    \phantomsection
    \label{tab:benchmark-scale-a}
    \textbf{\strut A. Engineering coverage}\par
    \vspace{0.25em}

    \begin{tabularx}{\linewidth}{@{}>{\raggedright\arraybackslash}Xr@{}}
      \toprule
      Engineering family & Tasks \\
      \midrule
\input{generated/task_family_compact_rows.tex} 
      \midrule
      Total & \BenchmarkTaskCount{} \\
      \bottomrule
    \end{tabularx}
  \end{minipage}
  \hfill
  \begin{minipage}[t]{0.46\linewidth}
    \phantomsection
    \label{tab:benchmark-scale-b}
    \textbf{\strut B. Task and verification scale}\par
    \vspace{0.25em}

    \begin{tabularx}{\linewidth}{@{}>{\raggedright\arraybackslash}Xrr@{}}
      \toprule
      Attribute & Median & Max. \\
      \midrule
\input{generated/benchmark_scale_rows.tex} 
      \bottomrule
    \end{tabularx}
  \end{minipage}
  \end{center}
\end{table}

\hyperref[tab:benchmark-scale-b]{Table~\ref*{tab:benchmark-scale}B} reports oracle-solution size for
the tasks: the median oracle solution changes 553 lines across seven files, with maxima of 6,077
lines and 35 files. The full distributions of oracle lines and files changed reveal smaller patches
not apparent from the summary statistics (Appendix Figure~\ref{fig:benchmark-scale-distributions}).
SWE-Serve's oracle solutions change more lines and files on average than those of SWE-bench
Verified, SWE-bench Pro, and DeepSWE, while its mean prompt is shorter than those of DeepSWE
and SWE-bench Pro (Appendix Figure~\ref{fig:benchmark-scope-comparison};
\citealp{huang2026deepswe}). The oracle comparison suggests broader repository-level
implementation scope, while the shorter prompts leave agents to determine how
to implement that work rather than follow prescriptive plans.

Before examining any model outcomes, we used only the task instruction and tests in the executable
verifier to assign each task three labels: runtime-domain breadth, persistent state, and concurrent
coordination. Runtime-domain breadth indicates how many of four domains the verifier tests:
request I/O, scheduling and request lifecycle, model execution, and KV-cache and resource
management. These domains were derived from recurring runtime responsibilities in four prominent
open-source inference systems: vLLM~\citep{kwon2023efficient}, SGLang~\citep{zheng2024sglang},
TensorRT-LLM~\citep{tensorrtllm2026architecture}, and
Triton~\citep{triton2026architecture}, as documented in Appendix
Table~\ref{tab:inference-taxonomy-components}.
Persistent state indicates that the verifier tests state maintained across operations or lifecycle
phases. Concurrent coordination indicates that the verifier tests coordination between overlapping
operations. Appendix Table~\ref{tab:task-characterization} reports these labels alongside each
task's family, hardware, and other verifier attributes.

\subsection{Construction and Qualification}
\label{sec:construction-qualification}

Construction had three stages: first, five discovery paths identified and screened 786
task-source records, retaining 203 source candidates; next, task construction produced 156 task
candidates; finally, qualification admitted 53 tasks (Appendix
Figure~\ref{fig:iclr-construction-funnel}). The discovery paths used four sources: two release-note
scans, a manual review of release materials and associated pull requests, maintainer-roadmap issues,
and direct SGLang-team suggestions. Next, task construction deduplicated the 203 source candidates
and created task candidates. Each task candidate comprised a task instruction, a designated base
commit, a sandboxed execution environment specification, an oracle solution, and hidden tests.
For multi-PR candidates, the designated base commit precedes the first source change.

At the final stage, 53 of the 156 task candidates were admitted to SWE-Serve, a
34.0\% admission rate. Admission required each candidate to run in its specified hardware environment and
satisfy the following executable controls: the no-op had to fail every F2P and pass every P2P test,
while the oracle had to pass all tests; for performance tasks, these controls were repeated across
H100 nodes to account for timing variability. Alongside these executable controls, every admitted
task underwent agent-assisted
adversarial probing using agent-created patches and trajectories to identify vulnerabilities in its
task instruction and verifier (see Appendix Table~\ref{tab:task-validation-coverage} for validation
coverage). This probing looked for valid alternatives rejected by the verifier,
incomplete or shortcut solutions that it accepted, unexercised requirements, and environment-induced
execution failures, paralleling Terminal-Bench 3's cheating-agent runs and instruction--verifier
alignment review~\citep{marten2026frontierbench}. Executable controls, agent-assisted adversarial
probing, and human adjudication jointly informed task revision and final-roster selection. Across
this process, 17 broad task candidates were replaced by narrower tasks, and 27 task candidates
required verifier changes (Appendix
Section~\ref{app:benchmark-construction-scope}). Upon admission, a benchmark
canary (a fixed string that supports future detection of exposure to released SWE-Serve artifacts)
was added to each released task configuration, following BIG-bench and
Terminal-Bench~\citep{srivastava2023bigbench,merrill2026terminalbench};
Appendix~\ref{app:release-canary} describes the canary's scope and limitations. Together, these controls
establish that each task is nontrivial at its base commit, oracle-solvable, reproducible on
declared hardware, and scored against its production correctness requirements.

\FloatBarrier

%% file: generated/task_family_compact_rows.tex
% Generated by scripts/paper/generate_dataset_tables.py; do not edit.
    Model and backend enablement & 12 \\
    Speculative and advanced decoding & 14 \\
    Kernels, quantization, and performance & 8 \\
    Caching and runtime state & 7 \\
    Distributed execution and scheduling & 4 \\
    Serving APIs and runtime correctness & 8 \\

%% file: generated/benchmark_scale_rows.tex
% Generated by scripts/paper/generate_dataset_tables.py; do not edit.
    Task-instruction words & 200 & 915 \\
    Oracle lines changed & 553 & 6{,}077 \\
    Oracle files changed & 7 & 35 \\
    Fail-to-pass (\FtwoP{}) tests & 7 & 246 \\
    Pass-to-pass (\PtwoP{}) tests & 10 & 2{,}034 \\

%% file: iclr2027_sections/04_experimental_setup.tex
\section{Experimental Setup}
\label{sec:experimental-setup}

All experiments use Harbor 0.13.1~\citep{harbor2026} to run the full 53-task benchmark in sandboxes
with CPU or GPU (H100) resources. The agent develops a solution until it finishes or
reaches 350 steps or 210 minutes, then the hidden verifier scores its patch; individual commands
time out after 120 seconds without ending the attempt. The 210-minute limit accommodates
SWE-Serve's long-running tasks: across nine effort-matched configurations, mean task wall-clock time is 43.3
vs. 34.3 minutes on DeepSWE, and 24.5\% vs. 3.5\% of task means fall in the one-to-four-hour bucket
(Appendix Figure~\ref{fig:runtime-comparison}). Furthermore, only 2.4\% (42/1,749) of attempts across the 11
top-per-model configurations ended as execution-limit failures: 15 at 210 minutes and 27 at 350 steps
(Appendix Figure~\ref{fig:failed-attempt-outcomes}).
We rerun a task only when the evaluation fails to produce a valid score because of an issue with the
model-serving route, runner, hardware, or verifier infrastructure. We do not rerun tasks because of
agent errors, budget exhaustion, test failures, or crashes caused by the agent-created patch.

We compare models using mini-SWE-agent v2.4.3, a model-agnostic harness developed by the team behind
SWE-bench and SWE-agent~\citep{jimenez2024swebench,yang2024sweagent,minisweagent2025}.
Within this harness, models use only Bash to inspect, edit, and test the repository. We then evaluate 11 models,
including frontier models: Claude Opus 5 and
Sonnet 5~\citep{anthropic2026opus5,anthropic2026sonnet5}; GPT-5.6 Sol, Luna, and
Terra~\citep{openai2026gpt56}; Kimi K3~\citep{moonshot2026kimik3}; DeepSeek V4 Flash
0731~\citep{deepseek2026v4,deepseek2026v4flash0731};
GLM-5.2~\citep{glmteam2026glm5,glmteam2026glm52}; Gemini 3.6
Flash~\citep{googledeepmind2026gemini36flash}; Laguna S 2.1~\citep{poolside2026lagunas21}; and
Inkling S~\citep{thinkingmachines2026inklingsmall}.
Appendix Table~\ref{tab:model-serving-configurations} lists serving, token-limit,
and harness configurations. To assess reasoning-effort sensitivity, we evaluate
two Claude and
three GPT-5.6 models at \texttt{low}, \texttt{medium}, \texttt{high}, \texttt{xhigh}, and
\texttt{max}: 31 model--effort configurations, each evaluated three times ($K=3$).

The primary leaderboard highlights each model's configuration with the highest mean pass@1,
the average of three run-level pass rates
(Section~\ref{sec:task-formulation-scoring}). Appendix~\ref{app:leaderboard-aggregation} details
pass@1 aggregation. We also measure pass@3, the task fraction solved at
least once, and \mbox{pass\textasciicircum{}3}, the fraction solved in all three
attempts~(\citealp{yao2025taubench}). Furthermore, to assess harness sensitivity, we compare two leading
model--effort configurations under mini-SWE-agent
and their native harnesses
(Section~\ref{sec:agent-harness-sensitivity}).

We enforce a closed-book evaluation because an open-book pilot confirmed task-specific
upstream retrieval by all three models evaluated
(Appendix Table~\ref{tab:appendix-open-book-retrieval}). This setting blocks public web
and upstream repositories. Only Hugging Face remains allowlisted because serving tasks
require model artifacts; task source repositories stay blocked,
and model communication remains separate. To assess closed-book integrity, we audit every trajectory
for prohibited retrieval. Every prohibited retrieval attempt was blocked, so no trial was invalidated on this basis.

\FloatBarrier

%% file: iclr2027_sections/05_results.tex
\section{Results}

\subsection{SWE-Serve Reveals a 40-Point Pass@1 Spread Across Evaluated Models}
\label{sec:model-performance}

Across 11 models, mean pass@1 under each model's best-performing configuration spans a
40-percentage-point range, from 75\% for Claude Opus 5 and GPT-5.6 Sol to 35\% for Inkling S.
Pass@1, however, captures only one dimension of
performance: among four configurations tied at 64\% pass@1, mean per-task cost varies by
7.6$\times$ (\$0.95--\$7.24) and wall-clock time by 3.9$\times$ (25.5--99.9 minutes), while output
tokens range from 61k to 98k and agent steps from 82 to 148. Thus, even among agents with identical
pass@1 under a shared harness, selection depends on how cost, token use, steps, and latency are
prioritized (Table~\ref{tab:primary-results}).

\begin{table}[!htbp]
  \caption{Primary leaderboard. For 11 models, we report the effort configuration with the
  highest observed mean pass@1 with mini-SWE-agent; rows are
  ordered by mean pass@1. The pass@1 entry reports the mean and 95\% confidence interval across
  three runs. The \mbox{pass\textasciicircum{}3} entry is the fraction of tasks solved in all three runs. The final four
  columns report means over task attempts.}
  \label{tab:primary-results}
  \begin{center}
  \normalsize
  \setlength{\tabcolsep}{2.2pt}
  \renewcommand{\arraystretch}{1.04}
  \begin{tabular}{@{}lcrrrrrr@{}}
    \toprule
    & & & & \multicolumn{4}{c}{Mean per task} \\
    \cmidrule(lr){5-8}
    Model & \shortstack{Reasoning\\effort} & \multicolumn{1}{c}{\shortstack{pass@1\\95\% CI}}
          & \multicolumn{1}{c}{\mbox{pass\textasciicircum{}3}}
          & \multicolumn{1}{c}{Cost} & \multicolumn{1}{c}{\shortstack{Output\\tokens}}
          & \multicolumn{1}{c}{\shortstack{Agent\\steps}}
          & \multicolumn{1}{c}{\shortstack{Wall-clock\\time (min)}} \\
    \midrule
\input{generated/iclr_primary_result_rows.tex} 
    \bottomrule
  \end{tabular}
  \end{center}
\end{table}

To assess reasoning-effort trade-offs, we evaluated each of five models spanning the Claude and
GPT-5.6 families at five effort settings, bringing the full evaluation to 31 model--effort
configurations (Appendix Table~\ref{tab:consistency-results}). These
effort sweeps show that resource trade-offs also arise within models as effort changes. Across the
five models, higher effort consistently increases cost, while performance gains vary substantially
across models and effort settings (Appendix Figure~\ref{fig:effort-cost-tradeoff}). For Claude Opus
5, moving from \texttt{high} to \texttt{max} effort raises pass@1 only modestly, from 74\% to 75\%,
while increasing mean cost by 76\% (\$9.89 to \$17.40 per task). This example shows that optimizing
for pass@1 alone can incur substantially higher resource costs for only a marginal performance gain.
Appendix Figures~\ref{fig:cost-vs-pass-rate}--\ref{fig:wall-time-vs-pass-rate} plot pass@1 against
cost, output-token use, and wall-clock time and show the observed Pareto frontier for each resource.

Pass@1 can also obscure differences in repeated-run reliability. For example, DeepSeek V4 Flash at
\texttt{max} effort and GPT-5.6 Luna at \texttt{xhigh} effort both achieve 55\% pass@1, yet DeepSeek
has higher coverage across retries (74\% versus 64\% pass@3), while Luna has higher consistency
(45\% versus 42\% \mbox{pass\textasciicircum{}3}). Table~\ref{tab:primary-results} reports
consistency, while Appendix Table~\ref{tab:consistency-results} reports both metrics for all 31
model--effort configurations.

Family-level results provide a complementary view of performance across the benchmark: pooled
pass@1 across the 11 best-performing model configurations ranges from 42.9\% for model and backend
enablement to 74.6\% for kernels, quantization, and performance (Appendix
Figure~\ref{fig:capability-by-family}). Taken together, SWE-Serve separates models across a 40-point
pass@1 range while revealing differences in resource use and repeated-run reliability among
configurations with similar pass@1.

%% file: generated/iclr_primary_result_rows.tex
% Generated by scripts/paper/generate_model_result_tables.py; do not edit.
    Claude Opus 5 & \texttt{max} & $75\% \pm 4\%$ & $70\%$ & \$17.40 & 122k & 134 & 57.5 \\
    GPT-5.6 Sol & \texttt{max} & $75\% \pm 6\%$ & $68\%$ & \$12.26 & 52k & 63 & 29.5 \\
    Claude Sonnet 5 & \texttt{xhigh} & $64\% \pm 4\%$ & $51\%$ & \$6.61 & 98k & 148 & 40.6 \\
    Kimi K3 & \texttt{max} & $64\% \pm 6\%$ & $51\%$ & \$7.24 & 79k & 117 & 99.9 \\
    GPT-5.6 Luna & \texttt{max} & $64\% \pm 4\%$ & $51\%$ & \$0.95 & 65k & 110 & 28.9 \\
    GPT-5.6 Terra & \texttt{max} & $64\% \pm 4\%$ & $55\%$ & \$5.06 & 61k & 82 & 25.5 \\
    DeepSeek V4 Flash (0731) & \texttt{max} & $55\% \pm 5\%$ & $42\%$ & \$0.69 & 99k & 173 & 36.4 \\
    GLM-5.2 & \texttt{max} & $48\% \pm 2\%$ & $34\%$ & \$2.10 & 53k & 101 & 34.0 \\
    Gemini 3.6 Flash & \texttt{high} & $48\% \pm 7\%$ & $30\%$ & \$4.84 & 99k & 92 & 37.3 \\
    Laguna S 2.1 & \texttt{max} & $46\% \pm 5\%$ & $26\%$ & \$0.33 & 147k & 184 & 56.9 \\
    Inkling S & \texttt{xhigh} & $35\% \pm 3\%$ & $25\%$ & \$0.44 & 36k & 96 & 17.6 \\

%% file: iclr2027_sections/06_production_contract.tex
\subsection{Local Correctness Does Not Ensure Production Correctness}

A patch may satisfy component
and integration tests yet fail when exercised end to end through a live model-serving path. SWE-Serve
measures this distinction with 276 model-serving end-to-end (E2E) tests across 19 tasks. These
tests launch a standalone serving process with the required model and assess its behavior through a
public request or evaluation interface. The model-serving E2E tests are production-grounded: 242
(87.7\%) are sourced or adapted from the SGLang repository; the remaining 34 were added by SWE-Serve
to cover accepted SGLang behavior not otherwise tested.

\begin{figure}[t]
  \centering
  \includegraphics[width=\linewidth]{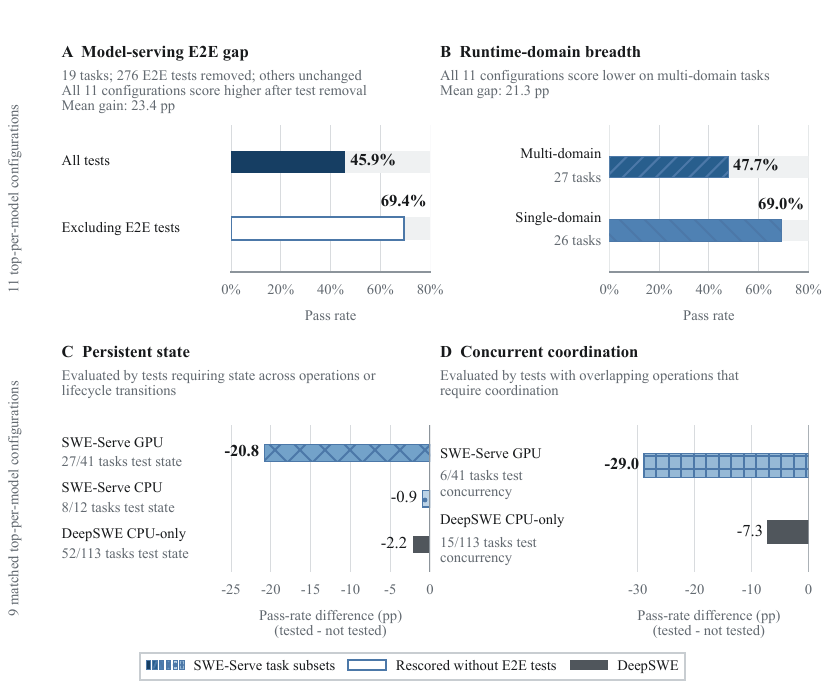}
  \caption{\textbf{Agentic implementations often fall short of production correctness.}
  \textbf{(A)} Rescoring the same patches without model-serving E2E tests raises pass rate from
  45.9\% to 69.4\%, a 23.4-percentage-point increase across 19 tasks.
  \textbf{(B)} Multi-runtime-domain tasks have pass rates 21.3 percentage points lower than
  single-runtime-domain tasks across 11 top-per-model configurations.
  \textbf{(C--D)} Across nine matched configurations, SWE-Serve GPU tasks whose tests require
  persistent state or concurrent coordination have pass rates 20.8 and 29.0 percentage points
  lower, respectively, than tasks not tested for these properties; gaps are smaller on
  DeepSWE~\citep{huang2026deepswe}.}
  \label{fig:production-contract}
\end{figure}

\textbf{Model-serving E2E gap.} To isolate model-serving E2E requirements, we remove these tests
from scoring while leaving the agent-created patches and all other tests unchanged. Across 19 tasks (627 patches), pass rate
increases from 45.9\% under the full verifier to 69.4\% when the E2E tests are removed from scoring,
a 23.4-percentage-point increase (Figure~\ref{fig:production-contract}A). Pass rate increases for all
11 top-per-model configurations (Appendix Figure~\ref{fig:production-contract-per-model}A).

\textbf{Matched test-removal control.} We test whether removing the same number of non-E2E tests
produces a comparable effect. Across 12 eligible tasks, we evaluate 10,000 randomized test-removal
pairs. Each pair excludes from scoring 38 E2E tests and 38 non-E2E tests matched within task and
F2P/P2P role, while holding all 396 agent-created patches fixed. E2E test removal produces 16.1
fail-to-pass transitions on average,
compared with 8.0 for matched non-E2E test removal, a 2.0$\times$ difference. E2E test removal
produces more fail-to-pass transitions in 92.2\% of the randomized pairs
(Appendix~\ref{app:production-contract-per-model}). The corresponding E2E-minus-control pass-rate gap
averages 2.1 points, with a 95\% mask-randomization interval of $-0.5$ to 4.3 points.

\textbf{SWE-Serve task example.} The Gemma 4 MoE core-serving task illustrates this pattern: 48.5\% of its agent-created patches
(16 of 33) passed every non-E2E test but failed at least one E2E test (Appendix
Table~\ref{tab:strict-e2e-task-decomposition}). These tests exercise loading the official checkpoint,
routing each token through the top eight of the model's 128 experts, text and image serving through
public interfaces, and ordered native batch generation with log
probabilities~\citep{gemmateam2026gemma4}.

\textbf{Sensitivity analyses.} Next, we assess robustness of the pass-rate increase
to task composition and tasks containing only E2E tests. First, resampling the 19
tasks 10,000 times yields a 95\% bootstrap confidence interval of 13.4--34.4 points for the
pass-rate increase (Appendix~\ref{app:production-contract-per-model}). Second, after excluding the
four tasks with no non-E2E tests, the pass-rate increase remains 22.8 points (Appendix
Table~\ref{tab:strict-e2e-task-decomposition}).

\textbf{Production inference task properties.} Beyond the model-serving E2E analysis, we examine
whether runtime-domain breadth, persistent state, and concurrent coordination are associated with
lower observed pass rates across the 11 top-per-model configurations. Runtime-domain breadth
measures coverage across four domains in the request-to-output path (Appendix
Table~\ref{tab:inference-taxonomy-components}); it does not cover infrastructure, deployment,
tooling, observability, or the full inference engineering stack. The 26 single-runtime-domain tasks
have a mean pass rate of 69.0\%, compared with 47.7\% for the 27 multi-runtime-domain tasks, an
average difference of 21.3 percentage points (Figure~\ref{fig:production-contract}B).
Multi-runtime-domain tasks have lower pass rates in every configuration
(Appendix~\ref{app:production-contract-per-model}).
Persistent state and concurrent coordination also show large performance differences among
SWE-Serve GPU tasks. Mean pass rate is 20.8 points lower for tasks requiring persistent state and 29.0
points lower for the six tasks that
explicitly test concurrent coordination (Figure~\ref{fig:production-contract}C--D). The
corresponding gaps are small on SWE-Serve CPU and DeepSWE tasks~\citep{huang2026deepswe}.

\textbf{Reasoning-effort robustness.} Across the GPT-5.6 model family, increasing reasoning effort
improves aggregate pass rates but does not consistently close the production correctness gaps,
which remain substantial even at maximum effort (Appendix Figure~\ref{fig:effort-dimension-robustness}).

\subsection{Sensitivity to Agent Harness}
\label{sec:agent-harness-sensitivity}

The primary leaderboard uses mini-SWE-agent, a model-agnostic harness~\citep{minisweagent2025},
to hold the agent scaffold constant across models. However, this may not reflect the performance
achievable with model-specific harnesses. We therefore re-evaluate the two best-performing
configurations, GPT-5.6 Sol and Claude Opus 5, both at maximum effort, using their model-specific harnesses,
Codex and Claude Code, respectively~\citep{openai2026codex,anthropic2026claudecode}.

\begin{figure}[!htbp]
  \centering
  \includegraphics[width=\linewidth]{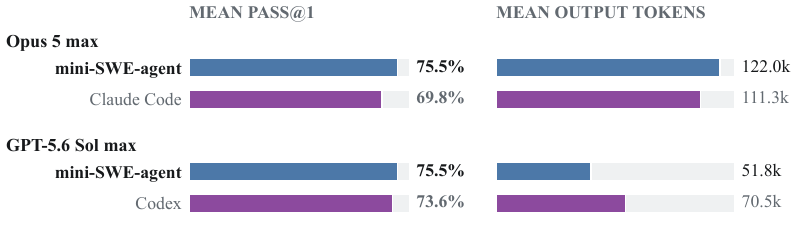}
  \caption{Model-specific harness sensitivity on 53 SWE-Serve tasks, with three runs per harness.
  mini-SWE-agent matches Codex for GPT-5.6 Sol and exceeds Claude Code for Opus 5.}
  \label{fig:native-harness-summary}
\end{figure}

Switching to model-specific harnesses does not improve mean pass@1 in either evaluated configuration.
mini-SWE-agent achieves 75.5\% for both configurations, compared with 73.6\% for GPT-5.6 Sol with
Codex and 69.8\% for Opus 5 with Claude Code (Figure~\ref{fig:native-harness-summary}). These results
support using mini-SWE-agent for the primary leaderboard, while showing that harness choice can
affect measured performance. The absence of an advantage for model-specific harnesses is consistent
with the finding reported for the DeepSWE benchmark~\citep{huang2026deepswe}.

Harness choice also affects resource trade-offs. For GPT-5.6 Sol, mini-SWE-agent achieves similar
pass@1 to Codex with fewer mean output tokens. For Opus 5, mini-SWE-agent achieves higher pass@1
than Claude Code with more mean output tokens. Appendix Figures~\ref{fig:code-agent-efficiency}
and~\ref{fig:code-agent-tokens} compare mean cost, wall-clock time, and token counts across harnesses.

%% file: iclr2027_sections/07_limitations_conclusion.tex
\section{Limitations}
\label{sec:limitations}

SWE-Serve provides an in-depth evaluation of production inference engineering. Extending executable
coverage beyond SGLang to other inference systems will help establish how broadly the findings
generalize. We restrict execution to CPU or a single H100 to reduce evaluation costs and broaden
access. This trade-off omits multi-GPU and multi-node tasks
involving parallelism, disaggregation, and distributed coordination. Additionally, executable
scoring measures functional, regression, integration, end-to-end, and targeted performance
requirements, but does not assess every dimension of upstream readiness. Patch size and change
dispersion complement these functional measurements as coarse indicators of implementation
complexity, but cannot establish maintainability, architectural fit, or replace maintainer review
(Appendix Figure~\ref{fig:appendix-scope-dumbbell} and
Table~\ref{tab:appendix-patch-scope}). Finally, closed-book execution and trajectory auditing
prevent agents from retrieving upstream solutions during evaluation, but cannot rule out prior
exposure to public SGLang code. To support future training-corpus decontamination after release,
each task includes a benchmark canary string (Appendix~\ref{app:release-canary}). A private held-out
task set would provide a complementary safeguard and remains a direction for future work.

\section{Conclusion}

We introduced SWE-Serve, a benchmark for evaluating agentic LLMs on challenging, repository-scale
production inference engineering tasks. We qualify its tasks with executable no-op and oracle
controls, agent-assisted adversarial probing, and targeted human review. Across SWE-Serve, agents
often satisfy local requirements without achieving production correctness. SWE-Serve quantifies
this gap directly: removing model-serving E2E tests from scoring while leaving the agent-created
patches and all other tests unchanged raises mean pass rate by 23.4 percentage points;
all 11 top-per-model configurations improve. SWE-Serve complements existing agentic benchmarks
by enabling the field to track whether future agents close the gap between completing tasks locally
and achieving production correctness.

%% file: iclr2027_sections/90_statements.tex
\subsubsection*{Acknowledgments}

We thank the SGLang maintainers and contributors for developing and maintaining the open-source
codebase on which SWE-Serve's benchmark tasks are based. We especially thank the maintainers for
their task suggestions, helpful pointers, and feedback throughout the development of this work.

%% file: iclr2027_sections/99_appendix.tex
\clearpage
% Top-align float-only appendix pages; retain horizontal centering and captions.
\makeatletter
\setlength{\@fptop}{0pt}
\setlength{\@dblfptop}{0pt}
\makeatother
\section{Appendix}

\subsubsection*{Appendix contents}

% The official ICLR style disables \addcontentsline. Use existing labels for
% linked titles, section numbers, and page numbers without modifying the style.
% Keep this list in sync when adding or moving appendix topics.
\begingroup
\newcommand{\appendixcontentsgroup}[1]{%
  \hyperref[#1]{\textbf{\ref*{#1}}} &
  \hyperref[#1]{\textbf{\nameref*{#1}}} &\pageref{#1} \\[5pt]}
\newcommand{\appendixcontentsentry}[1]{%
  \hspace*{0.75em}\hyperref[#1]{\ref*{#1}} &
  \hyperref[#1]{\nameref*{#1}} &\pageref{#1} \\[3pt]}
\noindent\begin{tabularx}{\linewidth}{@{}l@{\quad}Xr@{}}
  \multicolumn{2}{@{}l}{\textbf{Section}} & \textbf{Page} \\[5pt]
  \appendixcontentsgroup{app:construction-validation}
  \appendixcontentsentry{app:benchmark-scope-comparison}
  \appendixcontentsentry{app:inference-taxonomy}
  \appendixcontentsentry{app:model-serving-e2e-classification}
  \appendixcontentsentry{app:task-characterization}
  \appendixcontentsentry{app:benchmark-construction-scope}
  \appendixcontentsentry{app:release-canary}
  \appendixcontentsgroup{app:evaluation-methodology}
  \appendixcontentsentry{app:cross-benchmark-evaluation-runtime}
  \appendixcontentsentry{app:attempt-outcomes}
  \appendixcontentsentry{app:model-serving-configurations}
  \appendixcontentsentry{app:leaderboard-aggregation}
  \appendixcontentsentry{app:open-book-retrieval}
  \appendixcontentsgroup{app:additional-results}
  \appendixcontentsentry{app:model-effort-results}
  \appendixcontentsentry{app:performance-resource-tradeoffs}
  \appendixcontentsentry{app:task-family-characterization}
  \appendixcontentsgroup{app:correctness-robustness}
  \appendixcontentsentry{app:production-contract-per-model}
  \appendixcontentsentry{app:effort-dimension-robustness}
  \appendixcontentsgroup{app:agent-harness-sensitivity}
  \appendixcontentsgroup{app:scoring-limitations}
\end{tabularx}
\endgroup

\clearpage
\FloatBarrier
\subsection{Benchmark composition, construction, and qualification}
\label{app:construction-validation}

This section expands the composition, scope, construction, and qualification evidence in
Section~\ref{sec:benchmark}.

\subsubsection{Composition and scope}
\label{app:benchmark-scope-comparison}

\begin{figure}[H]
  \centering
  \includegraphics[width=0.90\linewidth]{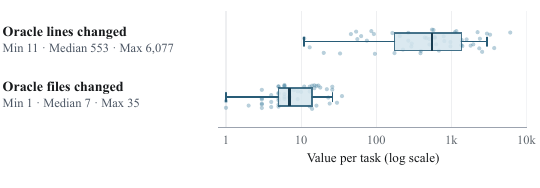}
  \caption{Oracle patch size and file count across SWE-Serve tasks. Dots show individual tasks;
  boxes show medians and interquartile ranges (IQRs). Whiskers reach the most extreme
  observations within $1.5\times$ IQR of the box edges, computed in original units.
  The value axis is logarithmic. Oracle lines changed counts additions plus deletions.}
  \label{fig:benchmark-scale-distributions}
\end{figure}

Figure~\ref{fig:benchmark-scale-distributions} complements the summaries in
\hyperref[tab:benchmark-scale-b]{Table~\ref*{tab:benchmark-scale}B} with the full distribution of oracle scope, including the number of lines and files changed.
The full distribution highlights that the median oracle solution changes 553 lines across seven
files, while individual tasks range from 11 to 6,077 lines and from one to 35 files, showing that
SWE-Serve includes both localized changes and broader implementations.

\begin{figure}[!t]
  \centering
  \includegraphics[width=\linewidth]{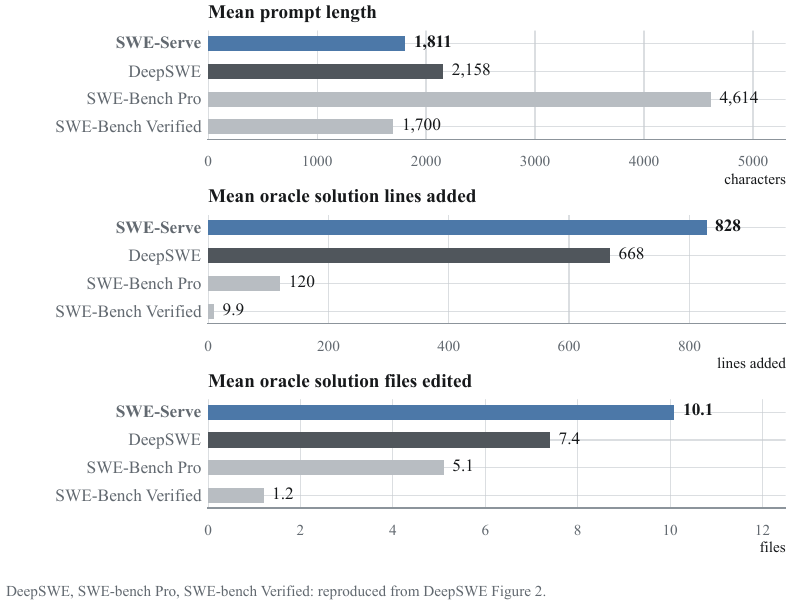}
  \caption{SWE-Serve's oracle solutions add more lines and modify more files on average than those of
  DeepSWE, SWE-bench Pro, and SWE-bench Verified, while its prompts are shorter on average than those
  of DeepSWE and SWE-bench Pro.
  Bars show mean prompt length in characters, lines added by the
  oracle solution, and files changed. SWE-Serve values cover the full benchmark; comparator values are
  reproduced from DeepSWE Figure~2~\citep{huang2026deepswe}.}
  \label{fig:benchmark-scope-comparison}
\end{figure}

Figure~\ref{fig:benchmark-scope-comparison} shows that SWE-Serve combines concise task instructions
with substantial oracle implementation scope. Its prompts are shorter on average than those of
DeepSWE and SWE-bench Pro, while its oracle solutions add more lines and modify more files on
average than those of DeepSWE, SWE-bench Pro, and SWE-bench Verified.
Instructions describe the desired behavior, and a separate executable verifier checks whether the
agent's solution satisfies the task's requirements. This allows prompts to remain concise while
leaving agents responsible for exploring the repository, planning the implementation, and producing
a patch that passes the tests.

% Let taxonomy prose fill the preceding page while Figure 5 floats to the top.
\subsubsection{Development of the runtime-domain taxonomy}
\label{app:inference-taxonomy}

To characterize the diversity of runtime responsibilities represented in SWE-Serve tasks, we
developed a runtime-domain taxonomy from documented components in four prominent open-source
inference systems: vLLM~\citep{kwon2023efficient}, SGLang~\citep{zheng2024sglang},
TensorRT-LLM~\citep{tensorrtllm2026architecture}, and Triton~\citep{triton2026architecture}.
We selected these systems for their popularity, as each official repository exceeded 10,000 GitHub
stars on September 1, 2026\footnote{GitHub repository snapshots:
\href{https://github.com/vllm-project/vllm}{vLLM} (90.7K),
\href{https://github.com/sgl-project/sglang}{SGLang} (33.0K),
\href{https://github.com/NVIDIA/TensorRT-LLM}{TensorRT-LLM} (14.5K), and
\href{https://github.com/triton-inference-server/server}{Triton} (11.0K).}, and for their differing
scopes: vLLM, SGLang, and TensorRT-LLM are LLM-serving runtime stacks, while Triton is a broader
inference server.

We examined the responsibilities of documented components along the path from receiving a request
to producing its output, then grouped components with corresponding responsibilities across systems
into four domains: \textbf{request handling and I/O}, covering input and output representation;
\textbf{scheduling and request lifecycle}, covering admission, queuing, batching, and request
progression; \textbf{model execution}, covering forward computation, decoding, attention, kernels,
and backend invocation; and \textbf{KV cache and runtime-resource management}, covering the
allocation and lifecycle of persistent inference-time state. These categories describe shared
responsibilities across systems whose component names and implementations differ. For example,
Triton's resource-management components handle sequence state and backend-managed resources,
whereas the LLM-serving systems include explicit KV-cache management.

% Keep the component table and its introduction after the scope figure.
\FloatBarrier

Table~\ref{tab:inference-taxonomy-components} maps these responsibilities to documented components,
with links to the supporting passages in pinned repository versions~\citep{vllm2026architecture,sglang2026repository,tensorrtllm2026architecture,triton2026architecture}.
We finalized the taxonomy and mapping before examining model
outcomes, then used the taxonomy to characterize the responsibilities exercised by each task's
executable tests.

\input{generated/inference_taxonomy_matrix}

% Let classification prose fill the page while the component table floats to the top.
\subsubsection{Model-serving end-to-end (E2E) test classification}
\label{app:model-serving-e2e-classification}

Motivated by SGLang maintainer feedback, we classified a test as model-serving E2E only if it met
all six requirements below:

\begin{enumerate}
  \item It uses a real pretrained checkpoint.
  \item It executes the actual production SGLang runtime without
  mocked or synthetic private substitutes.
  \item It drives the behavior under test through that runtime.
  \item It makes a caller-visible behavioral assertion.
  \item It launches a standalone serving process.
  \item It exercises that process through a public request or
  evaluation boundary, such as HTTP/OpenAI-compatible requests or an evaluation runner targeting
  the server.
\end{enumerate}

A task has model-serving E2E coverage when its executable verifier includes at least one such test
that contributes to the task's score.

\subsubsection{Task characterization}
\label{app:task-characterization}

Before examining any model outcomes, we defined four task characteristics: runtime-domain breadth
using the taxonomy in Appendix~\ref{app:inference-taxonomy}, persistent state and concurrent
coordination as defined in Section~\ref{sec:composition-scope}, and model-serving E2E coverage as
defined in Appendix~\ref{app:model-serving-e2e-classification}. Using these definitions, we generated
annotations for each characteristic across all SWE-Serve tasks. The annotation-creation process used only
each task's instruction and the tests in its executable verifier. We assigned a characteristic to a
task only when it was tested by the verifier, making it part of the task's executable correctness
requirements. An unassigned characteristic therefore means that the verifier does not establish that
requirement; it does not imply that an implementation that solves the task cannot choose to use
persistent state, coordinate concurrent operations, span additional runtime domains, or execute an
E2E serving path.

Beyond these four task characteristics, additional properties associated with each task include its
engineering family (summarized in
\hyperref[tab:benchmark-scale-a]{Table~\ref*{tab:benchmark-scale}A}), required execution hardware,
and the presence of performance gates.
Table~\ref{tab:task-characterization} reports the full characterization of each SWE-Serve task.

% Keep the component table ahead of the task-characterization longtable.
\FloatBarrier

\begingroup
\setlength{\tabcolsep}{2pt}
\renewcommand{\arraystretch}{1.00}
\setlength{\aboverulesep}{2pt}
\setlength{\belowrulesep}{2pt}
% Match the caption to the eight column widths plus their intercolumn padding.
\setlength{\LTcapwidth}{\dimexpr12.95cm+14\tabcolsep\relax}
% Keep the caption at manuscript size and separate it from the table by one line.
\makeatletter
\renewcommand{\LT@makecaption}[3]{%
  \LT@mcol\LT@cols c{\hbox to\z@{\hss
    \parbox[t]{\LTcapwidth}{\normalfont\normalsize
      #1{#2: }#3\par\vspace{\baselineskip}}%
    \hss}}}
\makeatother
\normalsize
\begin{longtable}{@{}>{\raggedright\arraybackslash}p{4.35cm}
                       >{\raggedright\arraybackslash}p{2.0cm}
                       >{\raggedright\arraybackslash}p{0.75cm}
                       >{\raggedright\arraybackslash}p{0.7cm}
                       >{\raggedright\arraybackslash}p{1.4cm}
                       >{\raggedright\arraybackslash}p{1.4cm}
                       >{\raggedright\arraybackslash}p{1.6cm}
                       >{\raggedright\arraybackslash}p{0.75cm}@{}}
  \caption{Task characterization of all SWE-Serve tasks. Engineering family identifies the type of
  engineering work. HW denotes the required execution hardware; all GPU tasks use an H100.
  E2E test denotes a model-serving end-to-end test, and Perf.\ gate abbreviates performance gate. In the model-serving E2E test,
  persistent state, concurrent behavior, and performance gate columns, a checkmark indicates a
  characteristic tested by the verifier as a correctness requirement. A dash means that the verifier
  does not establish this characteristic as a correctness requirement. The runtime-domains column indicates
  whether the verifier tests one (Single) or multiple (Multi) runtime domains. Runtime-domain codes
  are I=request handling and I/O, S=scheduling and request lifecycle, M=model execution, and
  K=key-value (KV) cache and runtime-resource management.}
  \label{tab:task-characterization}\\
  \toprule
  Task & Engineering\newline family & HW & E2E\newline test &
  Runtime\newline domains &
  Persistent\newline state & Concurrent\newline behavior &
  Perf.\newline gate \\
  \midrule
  \endfirsthead
  \multicolumn{8}{l}{\normalsize\itshape Table~\ref{tab:task-characterization}, continued.}\\
  \toprule
  Task & Engineering\newline family & HW & E2E\newline test &
  Runtime\newline domains &
  Persistent\newline state & Concurrent\newline behavior &
  Perf.\newline gate \\
  \midrule
  \endhead
  \midrule
  \multicolumn{8}{r}{\normalsize\itshape Continued on next page.}\\
  \endfoot
  \bottomrule
  \endlastfoot
\input{generated/task_characterization_rows.tex} 
\end{longtable}
\endgroup

\subsubsection{Source discovery, task construction, and qualification}
\label{app:benchmark-construction-scope}

\begin{figure}[p]
  \centering
  \includegraphics[width=\linewidth]{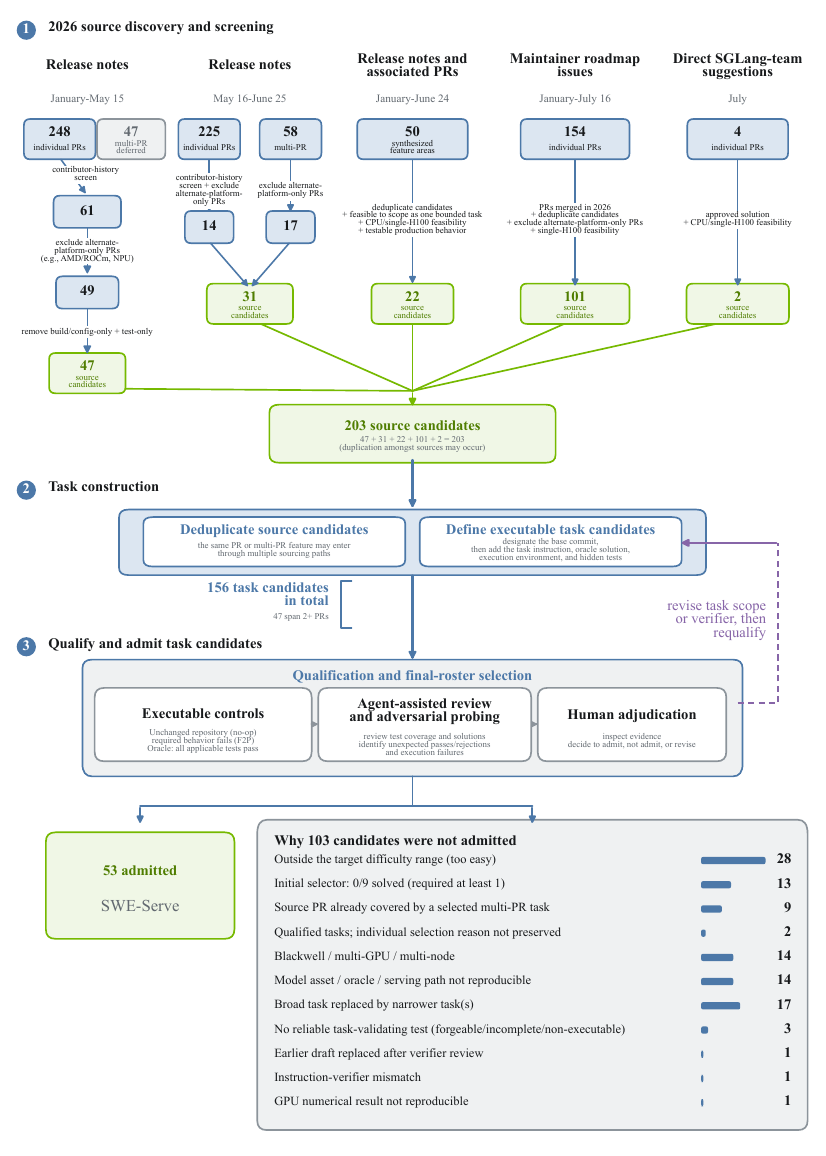}
  \caption{The 156 task candidates include every constructed or revised candidate version.
  Qualification may return candidates to construction for scope or verifier revision;
  human adjudication admits 53 task candidates and does not admit 103 candidates.}
  \label{fig:iclr-construction-funnel}
\end{figure}

The pipeline used to generate SWE-Serve comprised three stages: source discovery and screening,
task construction, and qualification and admission, as illustrated in
Figure~\ref{fig:iclr-construction-funnel}. Source discovery and screening used four source types to
identify and screen production inference engineering contributions made to the SGLang repository
since December 2025, yielding source candidates. Task construction deduplicated these source
candidates and turned them into executable task candidates. Qualification and admission used
executable controls, agent-assisted review and adversarial probing, and human adjudication to determine
whether each task candidate should be revised and requalified, excluded, or admitted. This process
produced the SWE-Serve benchmark.

% Let the source descriptions fill the preceding text page while the funnel
% occupies its own float page.

\paragraph{Source discovery and screening.}
\mbox{}\par

\subparagraph{Sources.}
We used four source types through five discovery and screening paths.

\textbf{Release notes.} We reviewed SGLang release-note bullets to extract individual PRs or groups
of related PRs. Bullets citing multiple PRs were treated as a multi-PR source. We conducted two
rounds of review, with the second incorporating newer release notes and removing candidates already
retained during the first round. Together, the two rounds covered release notes published from
January through June 25, 2026.

\textbf{Release notes and associated PRs.} We manually reviewed SGLang release notes from January
through June 24, 2026, and their associated PRs to identify broader features whose scope was not fully
captured by a single release-note bullet. This review connected related changes across bullets and
PRs to synthesize additional feature-level source candidates.

\textbf{Maintainer roadmap issues.} We reviewed maintainer-authored roadmap issues and extracted
PRs they linked to as individual PR source candidates. PRs that were part of the same feature were
subsequently combined during task construction.

\textbf{Direct SGLang-team suggestions.} Additionally, SGLang maintainers nominated four recent PRs
for review as potential benchmark contributions.

% Let screening criteria continue into the space before the funnel float page.
\subparagraph{Screening criteria.}
Each discovery path applied the screening criteria specified for that path in
Figure~\ref{fig:iclr-construction-funnel}.

The \textbf{contributor-history screen} selected individual PRs whose authors had both authored and
reviewed at least three PRs within a snapshot of the 500 most recently created SGLang PRs that had
been merged at the time. This served as a recent-activity heuristic rather than a measure of
contributors' entire contribution history. Additionally, authors needed to have submitted a review
within that snapshot with GitHub association \texttt{OWNER}, \texttt{MEMBER}, or
\texttt{COLLABORATOR}. This criterion applied to individual PRs in both rounds of release note review.

\textbf{Exclude alternate-platform-only PRs.} To support benchmark execution within a standardized
hardware configuration with CPUs and a single H100 GPU, we excluded contributions focused only on
hardware or software backends outside that configuration, such as AMD/ROCm and Ascend/NPU. Release
note review identified these contributions using PR titles and labels, while roadmap review
examined their implementation scope.

The \textbf{remove build/config-only + test-only screen} excluded candidates whose code changes
were limited to build files, configuration, or tests. This criterion applied during the first round
of release note review.

During manual feature review, we inspected the associated PR diffs to apply the deduplication,
bounded-task scope, hardware-feasibility, and testable-production-behavior criteria.
\textbf{Deduplication} was performed by assessing overlap with existing source candidates. Next,
\textbf{feasible to scope as one bounded task} required a feature to support a clearly defined
behavioral objective. Then \textbf{testable production behavior} required a way to deterministically
test that behavior, although the tests could be authored during task construction.

The \textbf{CPU or single H100 GPU feasibility criterion} assessed whether the intended behavior
could be exercised on CPU or a single H100 GPU, keeping benchmark execution within a standardized
hardware configuration. This criterion applied to manual feature review and direct suggestions.
Roadmap screening specifically required feasibility on a single H100 GPU.

For roadmap candidates, \textbf{PRs merged in 2026} restricted eligibility to changes merged from
January 1 through the July 16 screening cutoff. \textbf{Deduplication} also applied during roadmap
screening.

For direct suggestions, \textbf{approved solution} required each PR to have an accepted implementation
that could serve as the task's reference solution.

\subparagraph{Screening outcome.}
Across the five discovery and screening paths, we retained 47 source candidates from the first
round of release note review, 31 from the second round, 22 from manual review of release notes and
associated PRs, 101 from maintainer roadmap issues, and two from direct SGLang-team suggestions.
This yielded 203 source candidates for task construction.

\paragraph{Task construction.}
\mbox{}\par

Task construction converted 203 source candidates into 156 task-candidates, including revised
versions. This process first deduplicated overlapping source candidates and then, where feasible,
developed the remaining candidates into executable tasks. Each executable task-candidate required a
designated base commit, task instruction, oracle solution, reproducible execution environment, and
hidden tests. Only source candidates that met these requirements became task-candidates, and each
could include one or several related PRs.

During qualification, some task-candidates were returned to task construction to revise their scope
or verifier before being requalified.

\paragraph{Qualify and admit task candidates.}
\mbox{}\par

The final admission stage applied three qualification steps and final-roster selection to all
156 task-candidates. The qualification steps included executable controls, agent-assisted review and
adversarial probing, and human adjudication.
Qualification and final-roster selection resulted in 53 tasks being admitted to the SWE-Serve
benchmark, while 103 task-candidates were not admitted.

\subparagraph{Executable controls.}
The unchanged base commit served as the no-op control. It was
required to pass all pass-to-pass tests and fail all fail-to-pass tests. The oracle solution was
required to pass all applicable tests, including both pass-to-pass and fail-to-pass tests.

\subparagraph{Agent-assisted review and adversarial probing.}
Review assessed whether the tests covered the behavioral requirements stated in the task
instructions. Agent-generated solutions and execution traces were also used to identify incomplete
or shortcut solutions that unexpectedly passed the verifier, valid solutions that were incorrectly
rejected, and failures caused by the execution environment. These findings provided evidence for
human adjudication of whether candidates required revision, exclusion, or admission.

\subparagraph{Human adjudication.}
Human reviewers examined evidence from the executable controls and the agent-assisted review and
adversarial probing to determine whether each task-candidate should be admitted, excluded, or
revised and requalified.

\subparagraph{Task-candidate revision and requalification.}
Task scope and verifier revisions occurred
throughout construction and qualification, with revised candidates undergoing requalification. For
example, one qualification review of 54 task-candidates identified 27 requiring verifier changes.
Of these, 22 involved changes to the tests used for scoring, while the remaining five involved
verifier execution or how the verifier determined which source files to trust. Twenty-three of the
final 53 tasks incorporated verifier revisions from this review.

\subparagraph{Validation coverage.}
Table~\ref{tab:task-validation-coverage} summarizes validation coverage
for the admitted SWE-Serve tasks. All received no-op and oracle controls, test coverage and scoring review
(F2P/P2P), and task-specific agent-assisted adversarial review of candidate solutions. Nine tasks
also triggered additional review because no agent attempts had passed. Reviewers examined the instructions, tests, proposed
solutions, verifier outputs, and execution traces before interpreting these results as evidence of
difficulty. Four tasks that triggered this review later recorded successful attempts. The remaining
five had no successful attempts across the evaluated runs and received task-specific analyses of
the causes of failure.

\begin{table}[H]
  \caption{Validation coverage for all SWE-Serve tasks. ``Applicable'' in the last row refers
  to the nine tasks that triggered zero-pass review based on results available at the time of assessment.
  Four of these tasks subsequently recorded successful attempts.}
  \label{tab:task-validation-coverage}
  \begin{center}
  \normalsize
  \begin{tabular}{@{}p{0.72\linewidth}l@{}}
    \toprule
    Validation component & Coverage \\
    \midrule
\input{generated/task_validation_coverage_rows.tex} 
    \bottomrule
  \end{tabular}
  \end{center}
\end{table}

\subparagraph{Reasons for task-candidate non-admission.}
Figure~\ref{fig:iclr-construction-funnel} summarizes why
103 task-candidates were not admitted. Of these, 28 had high observed solve rates and fell outside
the target difficulty range. An early selection rule, applied only to a subset of task-candidates,
also excluded 13 candidates with no observed solves; this did not establish that those tasks were
invalid.

Task scope and overlap accounted for further exclusions: 17 candidates were too broad and were
replaced by narrower tasks, while nine had source PRs already covered by selected multi-PR tasks.

Execution constraints prevented admission of 14 candidates requiring hardware outside the
benchmark's standardized hardware configuration, such as Blackwell GPUs, multiple GPUs, or multiple
nodes. Another 14 candidates could not faithfully execute the accepted implementation or exercise
the intended production behavior within the benchmark environment, for example because of
incompatible dependencies.

Verification issues accounted for three candidates without reliable fail-to-pass correctness
checks, one with an instruction--verifier mismatch, and one with non-reproducible GPU numerical
results. Finally, two candidates passed qualification, but their individual reasons for
non-admission were not preserved.

\subsubsection{Release canary}
\label{app:release-canary}

Following the benchmark-canary convention used by BIG-bench and
Terminal-Bench~\citep{srivastava2023bigbench,merrill2026terminalbench},
SWE-Serve includes a fixed benchmark-level marker in every \texttt{task.toml} file and repeats it
in \texttt{README.md} and \texttt{CANARY.md}. The configuration parser ignores the marker because
it is a TOML comment, so it does not affect task execution. Upon public release, the marker will
help training-corpus builders filter SWE-Serve artifacts and provide a way to identify potential
exposure through otherwise unexplained verbatim reproduction.

\clearpage
\subsection{Experimental setup}
\label{app:evaluation-methodology}

Expanding on Section~\ref{sec:experimental-setup}, we provide additional details on execution
budgets, model configurations, score aggregation, and closed-book evaluation.

\subsubsection{Cross-benchmark evaluation wall-clock time}
\label{app:cross-benchmark-evaluation-runtime}

\begin{figure}[H]
  \centering
  \includegraphics[width=\linewidth]{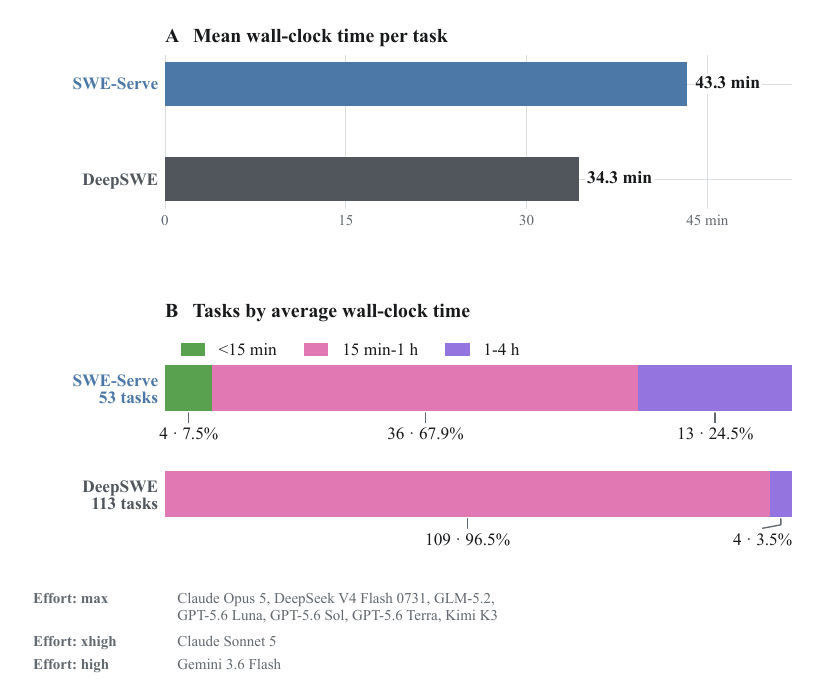}
  \caption{Average task wall-clock time for the nine of the 11 top-per-model configurations
  with effort-matched trajectories available in the DeepSWE release, all evaluated using
  mini-SWE-agent. Effort labels denote the resolved route settings. (A) The mean of the
  task-average wall-clock times. (B) Task counts and percentages of total tasks within each of
  three wall-clock time intervals. SWE-Serve averages are based on 53 tasks and 1,428 timed trials;
  DeepSWE averages are based on 113 tasks and 4,060 timed trials. The greater-than-four-hour
  bucket is omitted because neither benchmark has a task-average wall-clock time above four
  hours. This is consistent with both evaluations using agent-execution limits below four hours.}
  \label{fig:runtime-comparison}
\end{figure}

To contextualize the 210-minute agent execution limit in Section~\ref{sec:experimental-setup},
Figure~\ref{fig:runtime-comparison} compares task wall-clock times on SWE-Serve and
DeepSWE~\citep{huang2026deepswe} using the nine of the 11 top-per-model configurations in
Table~\ref{tab:primary-results} for which effort-matched trajectories are available in the
DeepSWE release. All configurations were evaluated with the mini-SWE-agent harness.
Within each benchmark, we first average wall-clock time across
all timed trials for each task, then average across tasks. SWE-Serve has a mean task wall-clock
time of 43.3 minutes, compared with 34.3 minutes for DeepSWE. Additionally, 24.5\% of SWE-Serve
tasks have average wall-clock times between one and four hours, compared with 3.5\% of DeepSWE
tasks. Differences in execution limits, endpoints, hardware, task mix, and verification make
this a descriptive comparison.

\FloatBarrier

\subsubsection{Attempt outcomes and execution limits}
\label{app:attempt-outcomes}

\begin{figure}[!htb]
  \centering
  \includegraphics[width=0.94\linewidth]{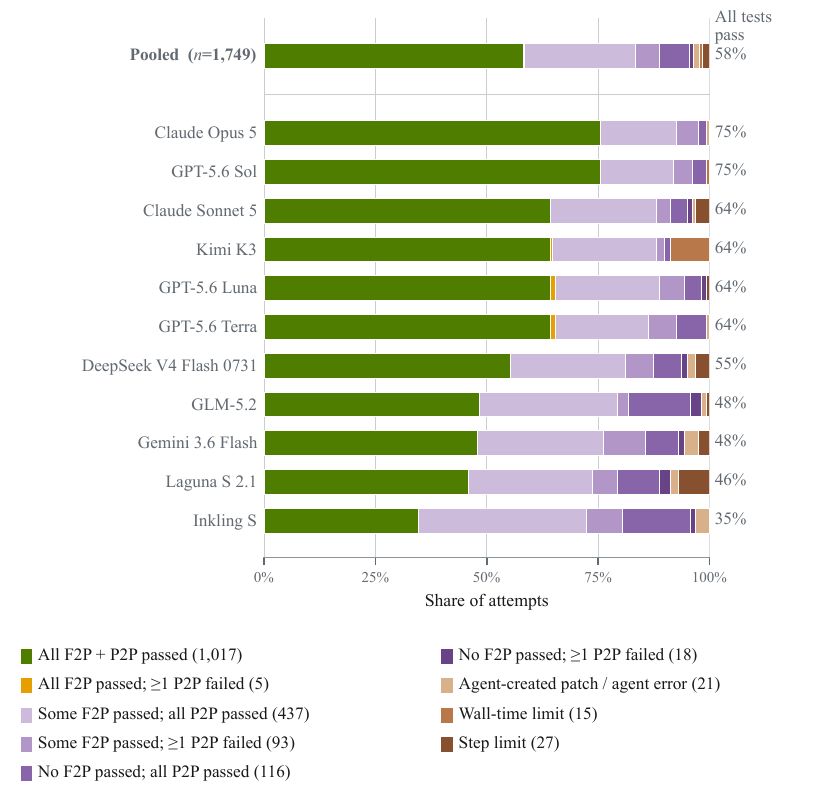}
  \caption{Outcomes for the 1,749 attempts from the 11 top-per-model configurations. The first
  six categories describe scored outcomes according to how many fail-to-pass (F2P) tests passed
  and whether all pass-to-pass (P2P) tests passed. ``Some F2P'' means strictly more than zero and
  strictly fewer than all F2P checks. Agent-created patch / agent error denotes errors caused by
  the agent-created patch or the agent that prevented the verifier from producing a pass/fail result. The wall-time and step-limit categories
  denote unsuccessful attempts reaching the 210-minute or 350-step agent-execution limit;
  patches that reached these limits and passed verification remain in the ``All F2P + P2P passed'' category. Model rows follow
  Table~\ref{tab:primary-results}, ordered by mean pass@1, with 159 attempts across three runs
  per model. The right column repeats each model's mean pass@1 from Table~\ref{tab:primary-results}.}
  \label{fig:failed-attempt-outcomes}
\end{figure}

Figure~\ref{fig:failed-attempt-outcomes} classifies all 1,749 attempts from the 11 top-per-model
configurations in Table~\ref{tab:primary-results} into nine mutually exclusive outcome categories.
Six categories distinguish whether all, some, or no F2P tests passed and whether all P2P tests passed.
The remaining three categories capture agent-created patch or agent errors, wall-time limits, and step limits.

Execution-limit failures account for 42 attempts (2.4\%): 15 at the 210-minute wall-time limit
and 27 at the 350-step limit. Wall-time limit failures therefore account for only 0.9\% of
attempts across the 11 top-per-model configurations. These limits constrain agent execution;
the resulting patch is still evaluated, and attempts that pass verification remain in the
``All F2P + P2P passed'' category. To contextualize this low wall-time failure rate, the DeepSWE
release's effort-matched results~\citep{huang2026deepswe} for nine of our 11 top-per-model
configurations contain 47 wall-time limit failures out of 4,060 attempts (1.2\%); three additional
timed-out attempts passed verification.

In 21 attempts (1.2\%), errors caused by the agent or its patch prevented the verifier from
producing a pass/fail result.

Among the remaining 669 scored failures, the most common outcome is passing some F2P checks
while preserving all P2P checks: 437 attempts (65.3\%). The proportion passing no F2P checks
while preserving all P2P checks also tends to be larger for configurations with lower pass@1.
Together, these failure patterns indicate that many unsuccessful patches preserve tested
existing behavior (P2P tests) while failing to satisfy all of the task's newly required behavior
(F2P tests).
\FloatBarrier

\subsubsection{Model, serving, and harness configurations}
\label{app:model-serving-configurations}

To support reproducibility, Table~\ref{tab:model-serving-configurations} records the model-serving
and evaluation settings for the 11 top-per-model configurations in Table~\ref{tab:primary-results},
together with the harness versions used in the native-harness sensitivity analysis
(Section~\ref{sec:agent-harness-sensitivity} and Appendix~\ref{app:agent-harness-sensitivity}).
These settings include reasoning effort, API interfaces, context and
output-token limits, and inference hardware for self-hosted models. All primary evaluations
use mini-SWE-agent v2.4.3.

The reasoning-effort analysis uses the same settings; however, it varies the effort setting
across \texttt{low}, \texttt{medium}, \texttt{high}, \texttt{xhigh}, and \texttt{max} for the two
Claude and three GPT-5.6 models.

\begin{table}[!htbp]
  \caption{Model-serving and evaluation settings for (a) provider-hosted and (b) self-hosted
  top-per-model configurations, and (c) the agent-harness sensitivity analysis.
  Context and output limits are reported
  separately in tokens; \emph{native} indicates that
  mini-SWE-agent imposed no token-limit constraints. All evaluations outside the harness-sensitivity
  analysis use mini-SWE-agent v2.4.3~\citep{minisweagent2025}. Part (c) applies only to the
  harness-sensitivity analysis in Section~\ref{sec:agent-harness-sensitivity} and
  Appendix~\ref{app:agent-harness-sensitivity}.}
  \label{tab:model-serving-configurations}
  \begin{center}
  \normalsize
  \setlength{\tabcolsep}{2pt}
  \renewcommand{\arraystretch}{1.05}

  \raggedright\textbf{(a) Provider-hosted top-per-model configurations}\par\vspace{2pt}
  \begin{tabularx}{\linewidth}{@{}>{\raggedright\arraybackslash}X>{\raggedright\arraybackslash}p{0.12\linewidth}>{\raggedright\arraybackslash}p{0.19\linewidth}>{\raggedright\arraybackslash}p{0.16\linewidth}>{\raggedright\arraybackslash}p{0.16\linewidth}@{}}
    \toprule
    \shortstack[l]{Model\\\strut} & \shortstack[l]{Reasoning\\effort} & \shortstack[l]{API\\interface} & \shortstack[l]{Context limit\\(tokens)} & \shortstack[l]{Output limit\\(tokens)} \\
    \midrule
    Claude Opus 5 & \texttt{max} & Messages API & 1,000,000 & 128,000 \\
    \addlinespace[3pt]
    Claude Sonnet 5 & \texttt{xhigh} & Messages API & 1,000,000 & 128,000 \\
    \addlinespace[3pt]
    GPT-5.6 Sol & \texttt{max} & Responses API & native & native \\
    \addlinespace[3pt]
    GPT-5.6 Luna & \texttt{max} & Responses API & native & native \\
    \addlinespace[3pt]
    GPT-5.6 Terra & \texttt{max} & Responses API & native & native \\
    \addlinespace[3pt]
    Gemini 3.6 Flash & \texttt{high} & Responses API & 1,048,576 & 65,536 \\
    \bottomrule
  \end{tabularx}

  \vspace{5pt}
  \raggedright\textbf{(b) Self-hosted top-per-model configurations}\par\vspace{2pt}
  \begin{tabularx}{\linewidth}{@{}>{\raggedright\arraybackslash}X>{\raggedright\arraybackslash}p{0.12\linewidth}>{\raggedright\arraybackslash}p{0.19\linewidth}>{\raggedright\arraybackslash}p{0.16\linewidth}>{\raggedright\arraybackslash}p{0.16\linewidth}@{}}
    \toprule
    \shortstack[l]{Model\\\strut} & \shortstack[l]{Reasoning\\effort} & \shortstack[l]{Model-serving\\hardware} & \shortstack[l]{Context limit\\(tokens)} & \shortstack[l]{Output limit\\(tokens)} \\
    \midrule
    Kimi K3 & \texttt{max} & 16 GB200 & 1,048,576 & 64,000 \\
    \addlinespace[3pt]
    GLM-5.2 (FP8) & \texttt{max} & 8 GB200 & 400,000 & 131,072 \\
    \addlinespace[3pt]
    DeepSeek V4 Flash 0731 & \texttt{max} & 4 GB200 & 1,048,576 & 384,000 \\
    \addlinespace[3pt]
    Laguna S 2.1 & \texttt{max} & 4 GB200 & 1,048,576 & 131,072 \\
    \addlinespace[3pt]
    Inkling S & \texttt{xhigh} & 4 GB200 & 262,144 & 64,000 \\
    \bottomrule
  \end{tabularx}

  \vspace{5pt}
  \raggedright\textbf{(c) Agent-harness sensitivity analysis}\par\vspace{2pt}
  \begin{tabularx}{\linewidth}{@{}>{\raggedright\arraybackslash}p{0.18\linewidth}>{\raggedright\arraybackslash}p{0.12\linewidth}>{\raggedright\arraybackslash}p{0.27\linewidth}>{\raggedright\arraybackslash}X@{}}
    \toprule
    \shortstack[l]{Model\\\strut} & \shortstack[l]{Reasoning\\effort} & \shortstack[l]{Shared\\harness} & \shortstack[l]{Model-specific\\harness} \\
    \midrule
    GPT-5.6 Sol & \texttt{max} & mini-SWE-agent v2.4.3
      & Codex CLI 0.144.5~\citep{openai2026codex} \\
    \addlinespace[3pt]
    Claude Opus 5 & \texttt{max} & mini-SWE-agent v2.4.3
      & Claude Code CLI 2.1.241~\citep{anthropic2026claudecode} \\
    \bottomrule
  \end{tabularx}

  \end{center}
\end{table}

\subsubsection{Leaderboard aggregation}
\label{app:leaderboard-aggregation}

Each of the 11 top-per-model configurations in Table~\ref{tab:primary-results} is evaluated in
three complete runs of the SWE-Serve benchmark. We calculate pass@1 as the mean of the
three run-level pass rates and report a 95\% confidence interval using the normal approximation,
$\bar r \mathbin{\pm} 1.96s_r/\sqrt{3}$, where $\bar r$ is the mean pass rate and $s_r$ is the
sample standard deviation of the three run-level pass rates. These intervals describe repeated-run
variability on the fixed benchmark. Additionally, both the pass@1 and 95\% CI percentages in
Table~\ref{tab:primary-results} are rounded to the nearest percentage point.

\subsubsection{Closed-book evaluation and retrieval calibration}
\label{app:open-book-retrieval}

An open-book pilot motivated the closed-book policy in Section~\ref{sec:experimental-setup}.
The pilot ran without network isolation. It evaluated the SWE-Serve benchmark using
mini-SWE-agent with GPT-5.6 Sol, Terra, and Luna under three reasoning-effort settings
(\texttt{xhigh}, \texttt{high}, and \texttt{medium}), with one complete benchmark run per
model--effort configuration. An \emph{external fetch} is a request to obtain information from a
public host outside the task's sandbox.
\emph{Confirmed out-of-band retrieval} requires recorded
output containing either the task's own source pull-request content or an upstream file modified
by its oracle patch. Table~\ref{tab:appendix-open-book-retrieval} reports the percentage of tasks
exhibiting each behavior for each model and effort setting. The \emph{Any model} column reports
the task-level union at that effort setting: a task is counted once if at least one of the three
model attempts exhibits the behavior specified by the row.

\begin{table}[!htbp]
  \caption{External fetches and confirmed out-of-band retrieval by agents in the open-book pilot.
  External fetches (requests to public hosts) and confirmed out-of-band retrieval (recorded output
  containing the task's source pull-request content or an upstream file modified by its oracle patch)
  are reported as percentages of SWE-Serve tasks for each model at three
  reasoning-effort levels: \texttt{xhigh}, \texttt{high}, and \texttt{medium}. The final \emph{Any model}
  column reports the percentage of tasks exhibiting each behavior in at least one of the three
  model attempts at the given effort level, counting each task once.}
  \label{tab:appendix-open-book-retrieval}
  \begin{center}
  \normalsize
  \begin{tabularx}{0.85\linewidth}{@{}>{\raggedright\arraybackslash}Xccc@{\hspace{18pt}}>{\bfseries}c@{}}
    \toprule
    Reasoning effort & Sol & Terra & Luna & \shortstack{Any\\model} \\
    \midrule
\input{generated/openbook_retrieval_by_model_rows.tex} 
    \bottomrule
  \end{tabularx}
  \end{center}
\end{table}

Out-of-band access is not isolated to one model or reasoning setting: all nine benchmark runs across
different model--effort configurations contain external fetches, and all three models exhibit
confirmed out-of-band retrieval. Incidence of both external fetches and out-of-band retrieval
generally increases with reasoning effort.

All other SWE-Serve evaluations enforced network isolation.
\clearpage
\subsection{Agentic performance and resource trade-offs}
\label{app:additional-results}

This section extends Section~\ref{sec:model-performance} with results for all 31 model--effort
configurations, resource trade-offs, and performance by engineering family.

\subsubsection{Performance across 31 model--effort configurations}
\label{app:model-effort-results}

Table~\ref{tab:consistency-results} extends the primary leaderboard in Table~\ref{tab:primary-results}
to show both performance and consistency across all 31 model--effort configurations. Alongside
mean pass@1, it reports pass@3 and \mbox{pass\textasciicircum{}3}; the gap between these measures
is the fraction of tasks solved in some but not all three attempts. The table includes five
reasoning-effort settings for each of the two Claude and three GPT-5.6 models, and one configuration
for each remaining model.

  \begin{table}[!htbp]
  \caption{Mean pass@1, pass@3, and \mbox{pass\textasciicircum{}3} for all 31 model--effort
  configurations, ordered by mean pass@1.
  Each configuration was evaluated with three complete runs of SWE-Serve. The pass@3 value is
  the fraction of tasks solved in at least one of the three runs; \mbox{pass\textasciicircum{}3}
  is the fraction solved in all three runs.}
  \label{tab:consistency-results}
  \begin{center}
  \normalsize
  \setlength{\tabcolsep}{8pt}
  \begin{tabular}{@{}lcrrr@{}}
    \toprule
    \multicolumn{2}{c}{Configuration} & & & \\
    \cmidrule(lr){1-2}
    Model & Reasoning effort & Mean pass@1 & pass@3 & \mbox{pass\textasciicircum{}3} \\
    \midrule
\input{generated/consistency_result_rows.tex} 
    \bottomrule
  \end{tabular}
  \end{center}
\end{table}

\clearpage
\subsubsection{Performance and resource trade-offs}
\label{app:performance-resource-tradeoffs}

\begin{figure}[!htbp]
  \centering
  \includegraphics[width=\linewidth]{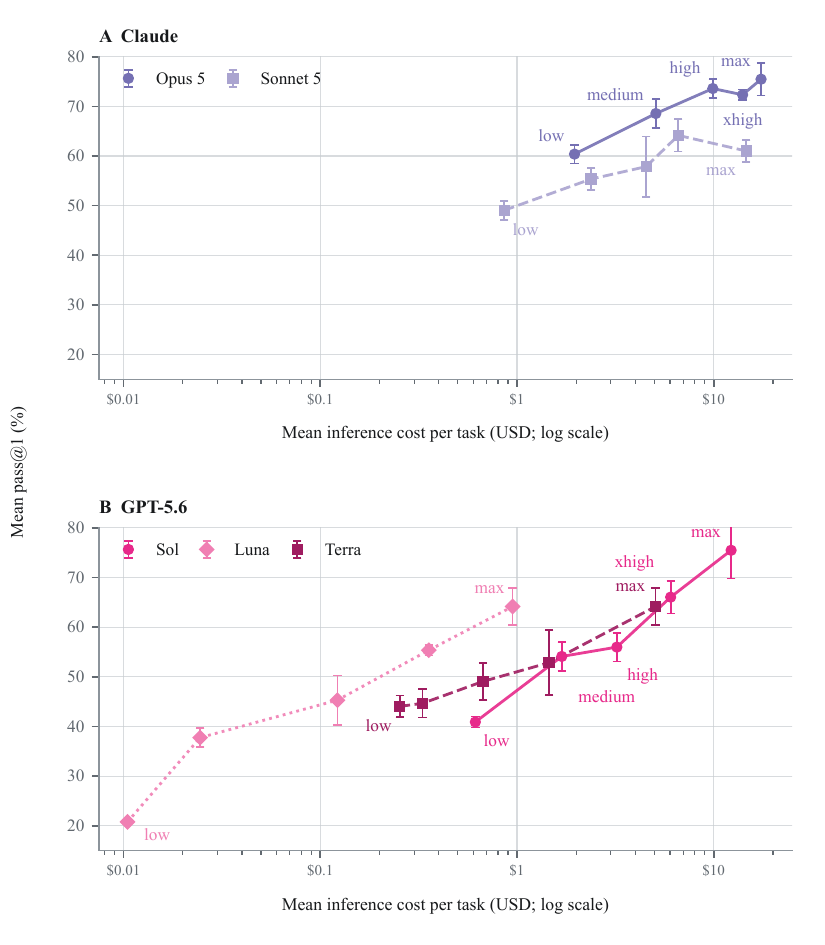}
  \caption{Mean pass@1 versus mean inference cost per task across reasoning-effort sweeps for
  (A) Claude Opus 5 and Claude Sonnet 5 and (B) GPT-5.6 Sol, Luna, and Terra. Axes are shared
  between (A) and (B); the cost axis is logarithmic. Vertical bars show one sample standard
  deviation across the three runs of SWE-Serve for each model--effort configuration. All five
  sweeps include \texttt{low}, \texttt{medium}, \texttt{high}, \texttt{xhigh}, and \texttt{max}
  reasoning effort. Higher reasoning effort is associated with higher mean inference cost,
  but not consistently with higher mean pass@1.}
  \label{fig:effort-cost-tradeoff}
\end{figure}

Figure~\ref{fig:effort-cost-tradeoff} compares mean pass@1 and inference cost across reasoning-effort
settings. Higher effort consistently increases cost, while performance gains vary across models
and effort settings.

\clearpage

\begin{figure}[htbp]
  \centering
  \includegraphics[width=\linewidth]{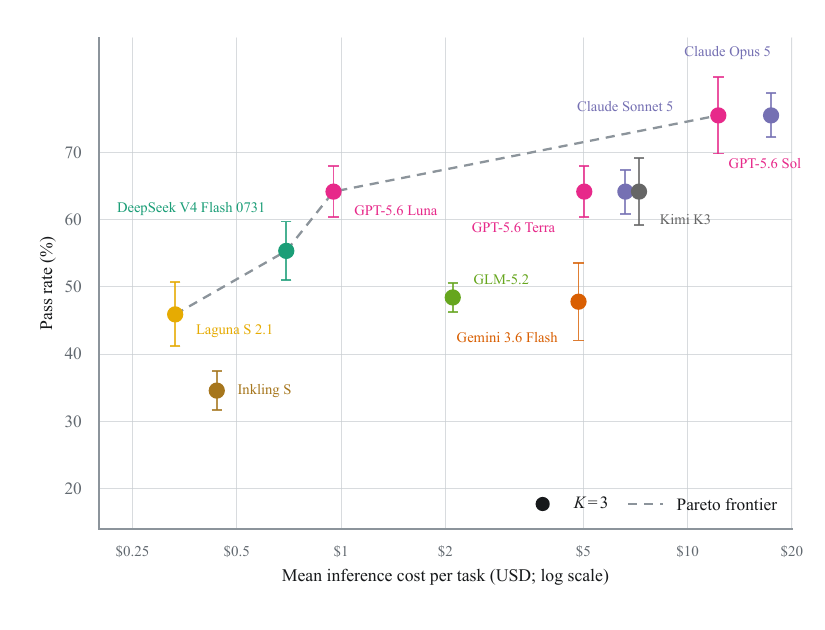}
  \caption{Mean pass@1 versus mean inference cost per task for the 11 top-per-model configurations.
  Vertical bars show one sample standard deviation across the three runs of SWE-Serve for each
  configuration. The cost axis is logarithmic. The dashed line connects the observed Pareto frontier
  for cost among these configurations. Higher mean inference cost is not consistently associated
  with higher mean pass@1, and configurations with similar mean pass@1 can differ in inference cost.}
  \label{fig:cost-vs-pass-rate}
\end{figure}

\begin{figure}[t]
  \centering
  \includegraphics[width=\linewidth]{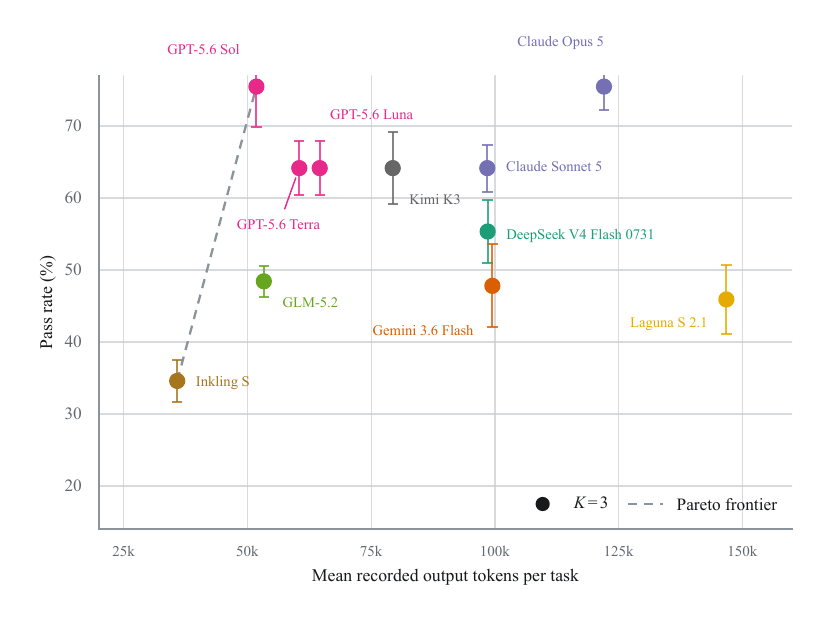}
  \caption{Mean pass@1 versus mean output tokens per task for the 11 top-per-model configurations.
  Vertical bars show one sample standard deviation across the three runs of SWE-Serve for each
  configuration. The dashed line connects the observed Pareto frontier for output-token use among
  these configurations. GPT-5.6 Terra's token count is based on 158 of 159 attempts as 1 value was
  missing and missing values are not extrapolated; all other configurations have complete token
  accounting. Higher mean output-token use is not consistently associated with higher pass@1:
  GPT-5.6 Sol matches Claude Opus 5's mean pass@1 while using fewer than half as many output tokens
  per task.}
  \label{fig:tokens-vs-pass-rate}
\end{figure}

\begin{figure}[t]
  \centering
  \includegraphics[width=\linewidth]{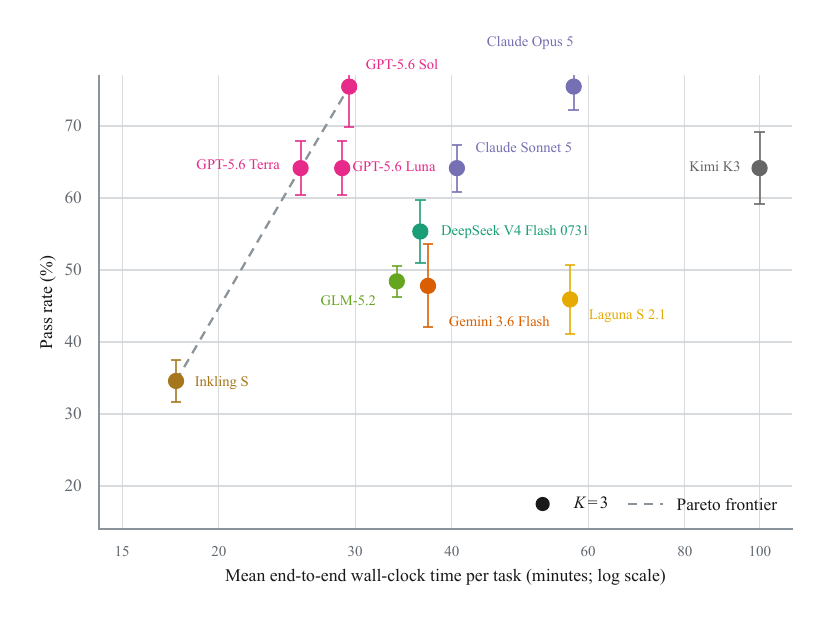}
  \caption{Mean pass@1 versus mean wall-clock time per task for the 11 top-per-model configurations,
  including setup, agent execution, and verification. Vertical bars show one sample standard
  deviation across the three runs of SWE-Serve for each configuration. The dashed line connects
  the observed Pareto frontier for wall-clock time among these configurations. Similar pass@1 can
  be associated with different latency: among the four configurations at approximately 64\% pass@1,
  mean wall-clock time varies by 3.9$\times$.}
  \label{fig:wall-time-vs-pass-rate}
\end{figure}

Figures~\ref{fig:cost-vs-pass-rate}--\ref{fig:wall-time-vs-pass-rate} compare the 11 top-per-model
configurations in Table~\ref{tab:primary-results}, plotting mean pass@1 against mean inference cost,
output-token use, and wall-clock time per task, respectively. Wall-clock time measures the
end-to-end duration of a task's attempt, including setup, agent execution, and verification.
Each plot shows the observed Pareto frontier among these selected configurations.
Even under a shared agent harness, configurations with similar pass@1 can differ substantially in
resource use, so configuration choice depends on the relative importance of performance, cost,
output-token use, and latency.

\FloatBarrier

\FloatBarrier

\subsubsection{Performance by inference engineering family}
\label{app:task-family-characterization}

\begin{figure}[H]
  \centering
  \includegraphics[width=\linewidth]{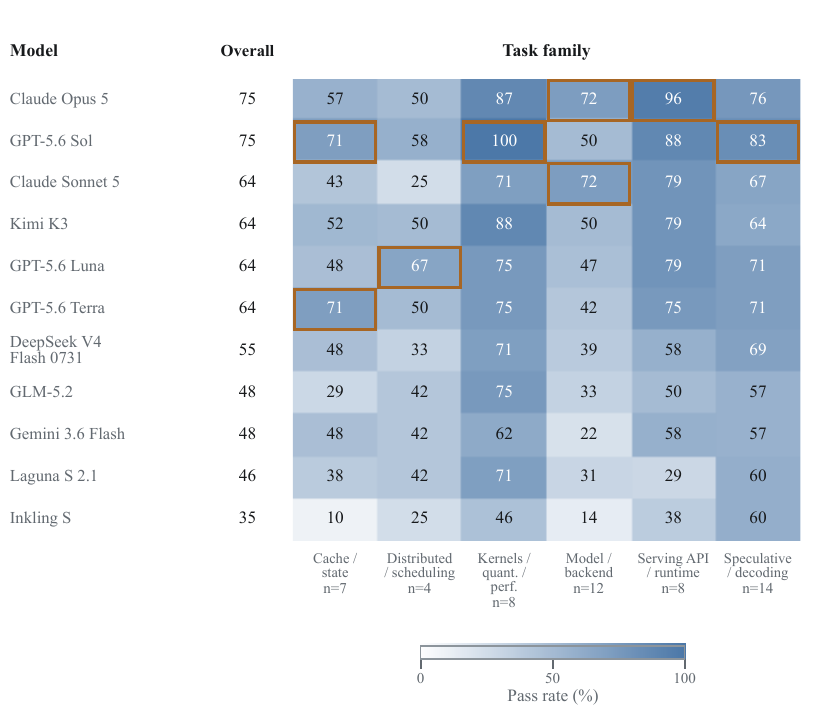}
  \caption{Engineering-family pass@1 for the 11 top-per-model configurations, ordered by mean
  pass@1 to match Table~\ref{tab:primary-results}. Scores are averaged across three SWE-Serve runs
  for each configuration. Task counts appear below the engineering-family column labels. Orange
  outlines mark the highest observed pass@1 in each family, including ties. The Overall column
  reports each configuration's mean pass@1 across the entire SWE-Serve benchmark. The Claude Opus 5
  and GPT-5.6 Sol configurations have the highest mean pass@1 across all tasks and collectively
  attain the highest mean pass@1 in five of the six task families, including ties. GPT-5.6 Luna has
  the highest mean pass@1 in distributed/scheduling tasks.}
  \label{fig:capability-by-family}
\end{figure}

Figure~\ref{fig:capability-by-family} reports pass@1 across the six inference engineering families
for the 11 top-per-model configurations. Averaged across these configurations, pass@1 ranges from
42.9\% for model and backend enablement to 74.6\% for kernels, quantization, and performance.
Claude Opus 5 and GPT-5.6 Sol, the two configurations with the highest overall mean pass@1,
collectively achieve the highest pass@1 in five of the six families. Claude Sonnet 5 and
GPT-5.6 Terra tie these leaders in model/backend enablement and caching/runtime state,
respectively, while GPT-5.6 Luna leads in distributed/scheduling tasks.

\clearpage
\subsection{Production correctness}
\label{app:correctness-robustness}

The following analyses expand the production correctness results in
Section~\hyperlink{subsection.5.2}{5.2} and Figure~\ref{fig:production-contract}, including task-level
and per-configuration breakdowns, sensitivity to which tests are removed and which tasks are
included, and comparisons of production correctness gaps across reasoning-effort settings.

\subsubsection{Local correctness does not ensure production correctness}
\label{app:production-contract-per-model}

\paragraph{Model-serving E2E gap.}
Table~\ref{tab:strict-e2e-task-decomposition} provides the task-level test counts and patch
outcomes behind the model-serving E2E gap in Figure~\ref{fig:production-contract}A. It covers
all 627 agent-created patches from the 11 top-per-model configurations on the
19 tasks that contain model-serving E2E tests. Of the 339 unsuccessful attempts under the full verifier,
removing model-serving E2E tests from scoring changes 147 patches from fail to pass, while 178
remain failures. These 147 fail-to-pass transitions out of 627 attempts produce the
23.4-percentage-point increase in pass rate across the 19 tasks.
The 14 attempts without a verifier pass/fail result remain counted as failures.
This analysis rescores recorded test results without rerunning the verifier.

\begin{table}[p]
  \caption{Task-level effects of removing model-serving E2E tests from scoring across 19 tasks
  with E2E tests. Each task has 33 attempts: three from each of the 11 top-per-model configurations.
  Test counts include only tests used for scoring. Outcome columns distinguish attempts that pass
  all tests, change from fail to pass after E2E removal, still fail after removal, or have no
  verifier pass/fail result. Attempts without a result remain failures. The first 15 tasks retain
  non-E2E tests after removal; the final four have no tests left in scoring.}
  \label{tab:strict-e2e-task-decomposition}
  \begin{center}
  \normalsize
  \def\UrlBreaks{\do\-}
  \setlength{\tabcolsep}{2.5pt}
  \renewcommand{\arraystretch}{1.02}
  \begin{tabularx}{\linewidth}{@{}>{\raggedright\arraybackslash}Xrr@{\hspace{9pt}}rrrr@{}}
    \toprule
    & \multicolumn{2}{c}{Test counts} & \multicolumn{4}{c@{}}{Number of attempts} \\
    \cmidrule(lr){2-3}\cmidrule(l){4-7}
    \shortstack[l]{19 tasks with\\model-serving E2E tests} & \shortstack{\# E2E\\tests} & \shortstack{\# Non-E2E\\tests}
      & \shortstack{Pass with\\all tests}
      & \shortstack{Fail $\to$ pass\\without E2E}
      & \shortstack{Still fail\\without E2E}
      & \shortstack{No pass/fail\\result} \\
    \midrule
\input{generated/strict_e2e_task_decomposition_rows.tex} 
    \bottomrule
  \end{tabularx}
  \end{center}
\end{table}

% Keep the full-size table and figure on float pages while the explanation flows.
\begin{figure}[p]
  \centering
  \includegraphics[width=0.90\linewidth]{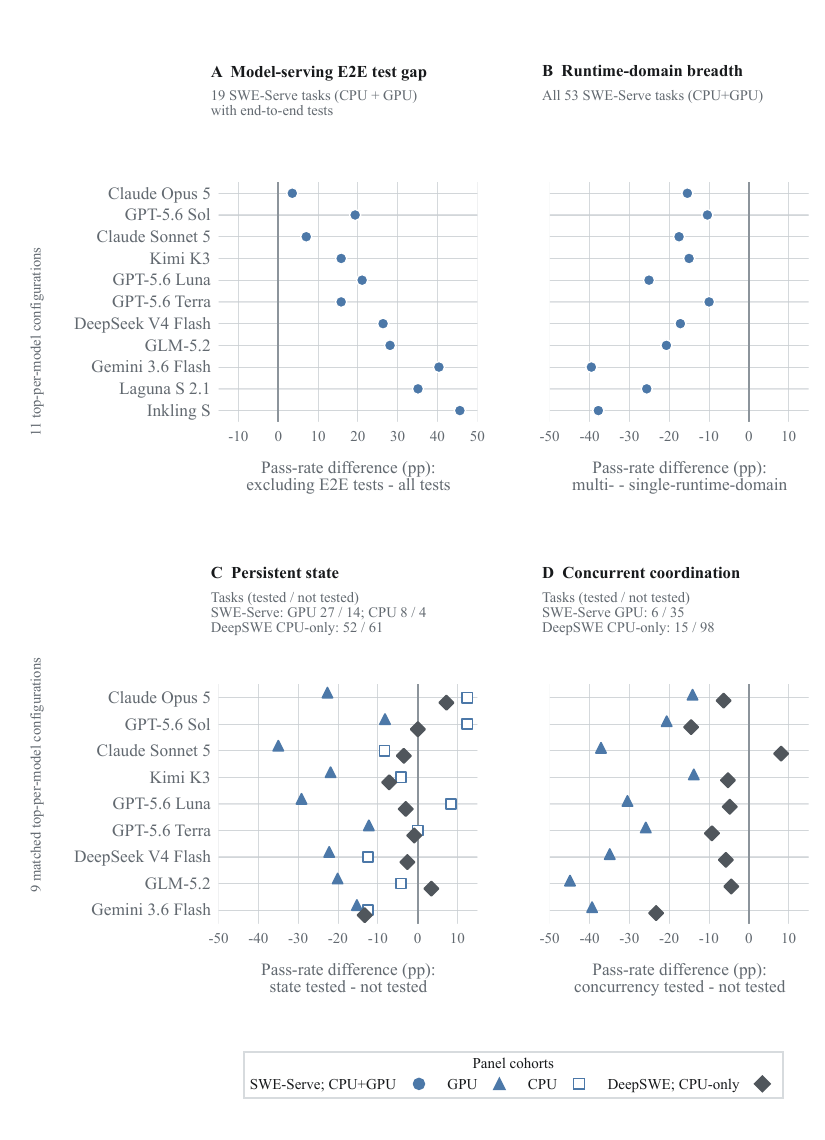}
  \caption{Production correctness gaps across top-per-model configurations.
  (A) Rescoring the same patches without model-serving E2E tests raises pass rate by
  3.5--45.6 percentage points across the 11 top-per-model configurations.
  (B) Multi-runtime-domain tasks have pass rates 10.0--39.6 percentage points lower than
  single-runtime-domain tasks across the 11 top-per-model configurations.
  (C--D) Across nine configurations with effort-matched trajectories available in the DeepSWE
  release, SWE-Serve GPU tasks tested for persistent state or concurrent coordination have pass
  rates 8.2--35.0 and 13.8--44.9 percentage points lower, respectively, than tasks not tested for
  these properties. CPU comparisons generally show smaller differences and vary in direction:
  panel C includes SWE-Serve CPU and DeepSWE, while panel D includes DeepSWE only~\citep{huang2026deepswe}.
  Panel D omits SWE-Serve CPU tasks because none are tested for concurrent coordination.
  Positive values in A indicate increased pass rates after removing E2E tests; negative values
  in B--D indicate lower pass rates on tasks with the specified production property.
  Configurations are ordered by overall mean pass@1 to match Table~\ref{tab:primary-results}.}
  \label{fig:production-contract-per-model}
\end{figure}

Figure~\ref{fig:production-contract-per-model}A breaks down the aggregate model-serving E2E gap
across the 11 top-per-model configurations in Figure~\ref{fig:production-contract}A by configuration.
For each of the configurations, it reports the increase in pass rate when the same patches on the
19 tasks are rescored without model-serving E2E tests. Pass rate increases for every configuration,
and the magnitude varies from 3.5 to 45.6 percentage points.

\paragraph{Matched test-removal control.}
To test whether removing equal numbers of non-E2E tests produces a comparable effect to removing model-serving E2E tests, we compare both conditions on 12 eligible tasks and their 396 patches. These 12 tasks are eligible
because we can remove matching numbers of E2E and non-E2E tests within the same F2P/P2P role.
Of the 19 tasks with E2E tests, four have no non-E2E tests, and three others have both test types
but never within the same role. Those seven tasks are therefore excluded.

For each task, we compare two versions of its score: one with some E2E tests excluded and another
with the same number of non-E2E tests excluded. We match the removal counts separately for F2P
and P2P tests, using the smaller available count. For example, if a task has two E2E F2P tests and
five non-E2E F2P tests, we exclude both E2E tests in the first condition and two randomly selected
non-E2E tests in the second. We apply the same rule to P2P tests. Across all 12 tasks, each scoring
condition excludes 38 tests: 19 F2P and 19 P2P.

We repeat the comparison 10,000 times, sampling tests without replacement separately for each
task, test type, and F2P/P2P role, using NumPy's PCG64 random-number generator with seed
20260826~\citep{harris2020numpy}. The tasks, patches, and removal counts remain fixed; only the
excluded tests vary.

\paragraph{Sensitivity to task composition.}
To assess how the E2E pass-rate increase depends on which tasks are included, we resample the 19 tasks
with model-serving E2E tests with replacement 10,000 times. We use NumPy's PCG64 random-number
generator with a fixed seed of 20260825 for reproducibility~\citep{harris2020numpy}. Each sampled task
retains all 33 agent-created patches, comprising the 11 top-per-model configurations and their three
attempts per task, so neither models nor attempts are independently resampled. We report a 95\% percentile bootstrap confidence interval.

\paragraph{Sensitivity to tasks containing only E2E tests.}
To assess whether the E2E pass-rate increase persists when other tests remain in scoring, we exclude
the four tasks containing only E2E tests (Table~\ref{tab:strict-e2e-task-decomposition}). Across the
remaining 15 tasks, pass rate still rises from 191/495 (38.6\%) to 304/495 (61.4\%), yielding a
22.8-percentage-point increase, corresponding to 113 fail-to-pass transitions.

\paragraph{Production inference task properties.}
Figure~\ref{fig:production-contract-per-model}B--D breaks down the aggregate production correctness
gaps in Figure~\ref{fig:production-contract}B--D by model configuration. The runtime-domain gap is
consistent across all 11 top-per-model configurations: pass rates are 10.0--39.6 percentage points
lower on tasks spanning multiple runtime domains than on single-domain tasks
(Figure~\ref{fig:production-contract-per-model}B).

\FloatBarrier

For persistent state and concurrent coordination, Figure~\ref{fig:production-contract-per-model}C--D
shows results for each of the same nine top-per-model configurations used in
Figure~\ref{fig:production-contract}C--D, evaluated at matching reasoning-effort settings on SWE-Serve
and DeepSWE~\citep{huang2026deepswe}. SWE-Serve GPU tasks testing persistent state or concurrent
coordination have lower pass rates than tasks where the corresponding property is not tested,
across all nine configurations. The corresponding differences on SWE-Serve CPU and DeepSWE are
generally smaller in magnitude and are not consistently negative across configurations.

\FloatBarrier

\subsubsection{Reasoning-effort robustness of production correctness gaps}
\label{app:effort-dimension-robustness}

\begin{figure}[!htb]
  \centering
  \includegraphics[width=0.85\linewidth]{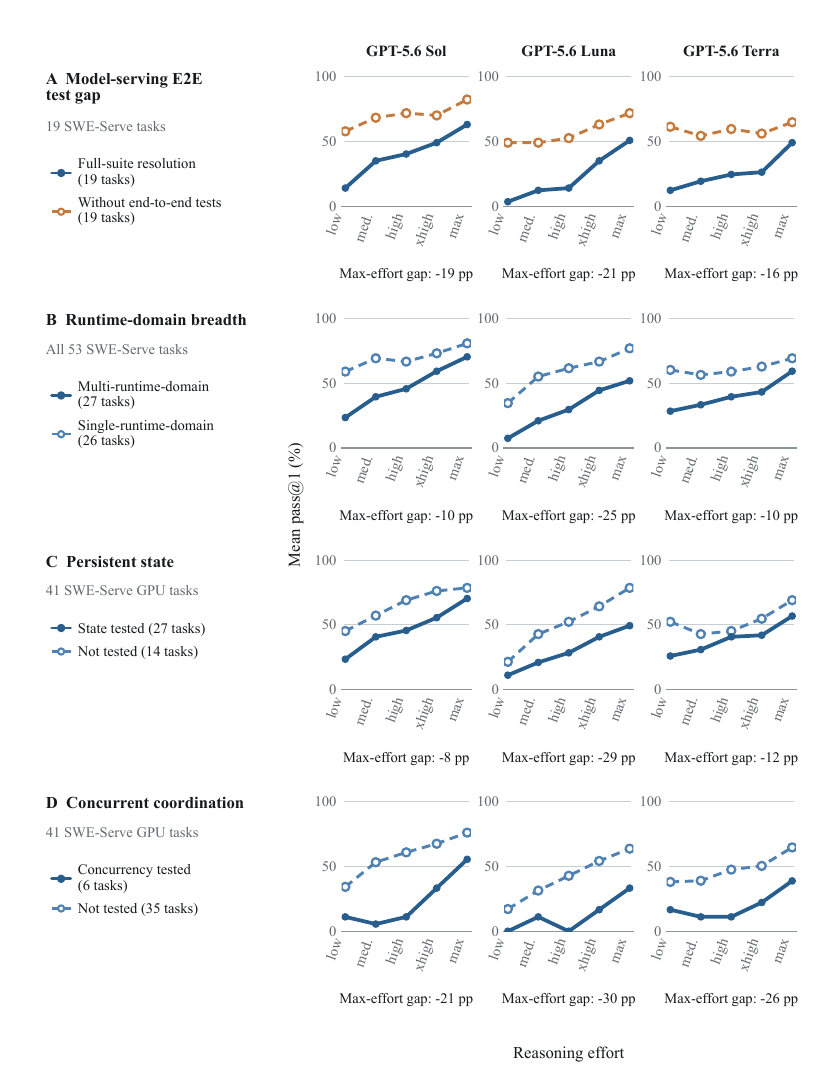}
  \caption{Higher reasoning effort narrows but does not eliminate the model-serving E2E gap across
  GPT-5.6 Sol, Luna, and Terra; task-property gaps also persist at maximum effort.
  Panels~A--D show the model-serving E2E test gap, runtime-domain breadth,
  persistent state, and concurrent coordination across five reasoning-effort settings, with three
  attempts per task for each configuration. Dark blue solid lines with filled markers show
  full-suite pass rates (A), multi-runtime-domain tasks (B), and tasks tested for the specified
  property (C--D). Lighter blue dashed lines with hollow markers show single-runtime-domain
  tasks (B) or tasks not tested for that property (C--D). Orange dashed lines in A show pass
  rates after removing model-serving E2E tests from scoring while keeping the same patches.
  Each ``Max-effort gap'' is the dark blue rate minus its comparison rate at \texttt{max} reasoning effort,
  in percentage points.}
  \label{fig:effort-dimension-robustness}
\end{figure}

Figure~\ref{fig:effort-dimension-robustness} examines whether higher reasoning effort narrows
production correctness gaps for GPT-5.6 Sol, Luna, and Terra across five settings: \texttt{low},
\texttt{medium}, \texttt{high}, \texttt{xhigh}, and \texttt{max}. It assesses all four production
correctness gaps using the same task groups as Figures~\ref{fig:production-contract}
and~\ref{fig:production-contract-per-model}. The E2E comparison rescores the same patches with
and without E2E tests; the other three comparisons measure pass-rate differences between task groups.

All four gaps remain at \texttt{max} reasoning effort. At this setting, the model-serving E2E gap is 16--21
percentage points, and pass rates on multi-runtime-domain tasks are 10--25 points below those
on single-domain tasks. Also at \texttt{max} reasoning effort, SWE-Serve GPU tasks testing persistent state
or concurrent coordination have pass rates 8--29 and 21--30 points lower, respectively, than
tasks where the corresponding property is not tested.

The response to increasing effort differs across the four comparisons. From \texttt{low} to \texttt{max}
reasoning effort, the E2E and runtime-domain gaps narrow for all three models. The persistent-state gap
narrows for Sol and Terra but widens for Luna, while the concurrent-coordination gap narrows
for Sol but widens for Luna and Terra. Even where gaps narrow overall, they do not always
shrink at each successive effort setting.

\clearpage

\subsection{Sensitivity to agent harness}
\label{app:agent-harness-sensitivity}

\begin{figure}[H]
  \centering
  \includegraphics[width=0.80\linewidth]{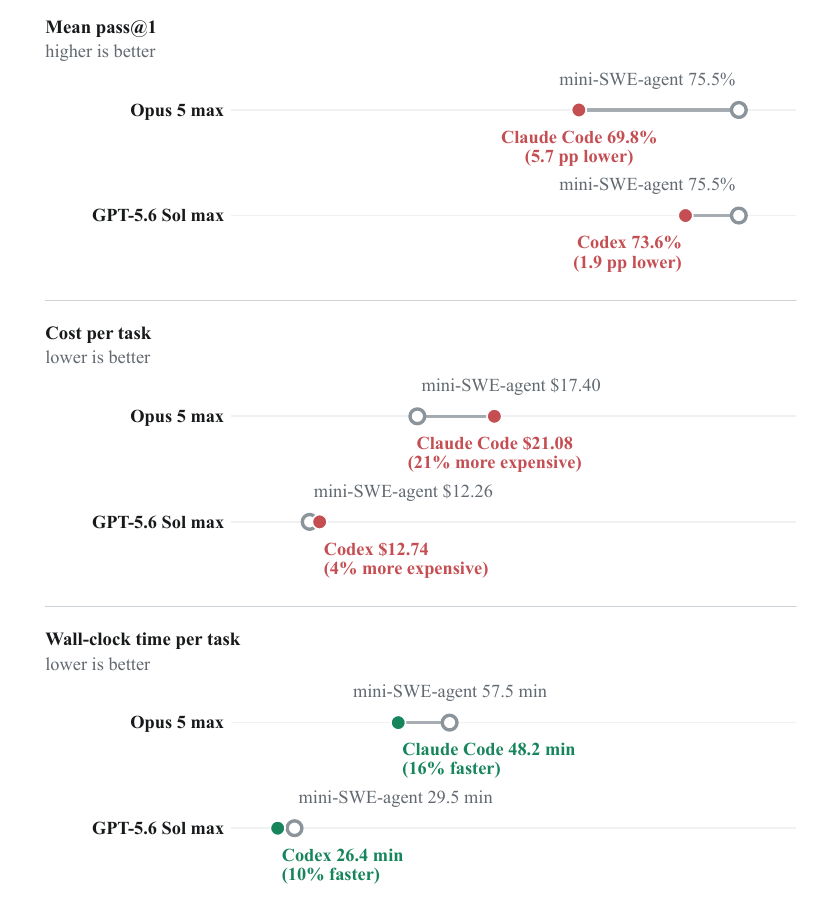}
  \caption{Comparing mean pass@1, cost, and wall-clock time for the two best-performing
  configurations, GPT-5.6 Sol and Claude Opus 5, when using their model-specific harness compared
  to the mini-SWE-agent harness. Each data point is based on three complete runs of SWE-Serve.
  Direct labels report each model-specific harness's change relative to mini-SWE-agent.
  Hollow points show the means for mini-SWE-agent, and filled points show the means for the
  model-specific harnesses. Costs for mini-SWE-agent and Claude Code are exact recorded endpoint
  costs, while Codex costs are estimated with approximately 1--2\% precision. For both evaluated
  configurations, mean wall-clock time is lower, inference cost is higher, and observed mean pass@1 is lower
  with the model-specific harness than with mini-SWE-agent.}
  \label{fig:code-agent-efficiency}
\end{figure}

\begin{figure}[H]
  \centering
  \includegraphics[width=0.80\linewidth]{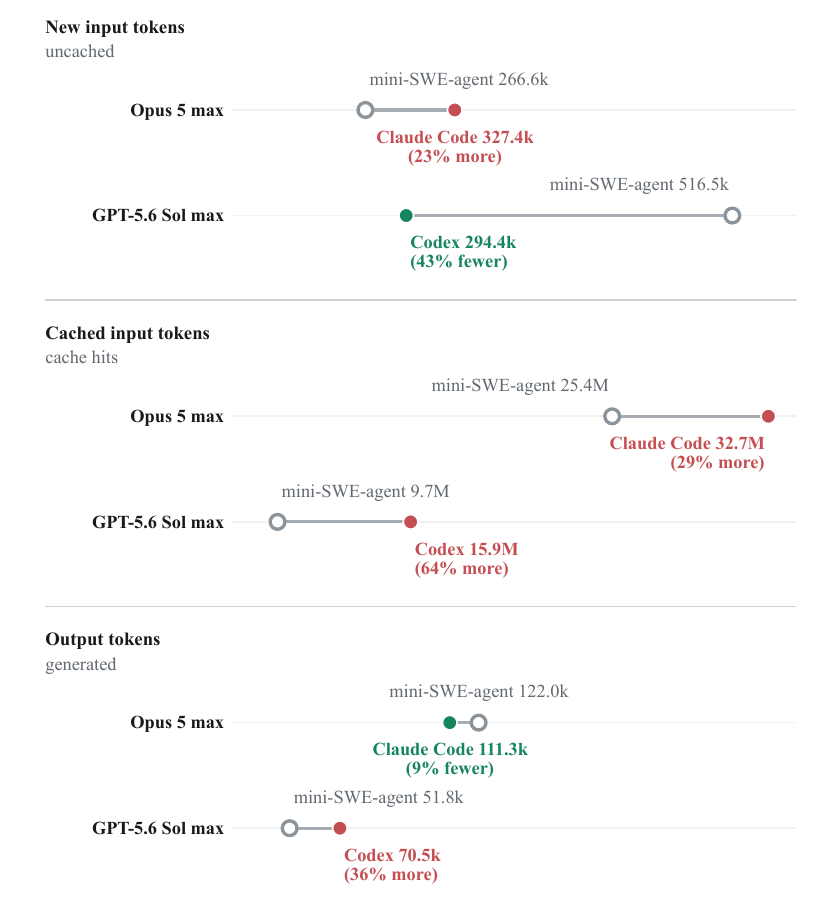}
  \caption{Mean per-task token use for GPT-5.6 Sol and Claude Opus 5, the two best-performing
  configurations, comparing their model-specific harnesses with mini-SWE-agent. Token use is
  separated into uncached input, cached input, and generated output tokens. Direct labels report
  each model-specific harness's relative change from mini-SWE-agent. Hollow points show means
  for mini-SWE-agent, and filled points show means for the model-specific harnesses. Relative to
  mini-SWE-agent, both model-specific harnesses have higher mean cached input-token use, but their
  other token-use changes differ: Codex has lower mean uncached input-token use and higher mean
  output-token use, while Claude Code shows the opposite pattern.}
  \label{fig:code-agent-tokens}
\end{figure}

This section expands the agent-harness sensitivity analysis in Section~\ref{sec:agent-harness-sensitivity}
with details of the analysis and how agent scaffold choice affects resource use. We compare
mini-SWE-agent with the model-specific harnesses for GPT-5.6 Sol and Claude Opus 5, the two
best-performing configurations, both at maximum reasoning effort.
Figure~\ref{fig:code-agent-efficiency} reports mean inference cost and wall-clock time per task,
while Figure~\ref{fig:code-agent-tokens} breaks down mean token use into uncached input, cached
input, and output tokens.

We evaluated GPT-5.6 Sol and Claude Opus 5, both at maximum reasoning effort, with three complete
runs of SWE-Serve under their model-specific harnesses: Codex CLI 0.144.5~\citep{openai2026codex}
for GPT-5.6 Sol and Claude Code CLI 2.1.241~\citep{anthropic2026claudecode} for Claude Opus 5.
Using these harnesses changes the full agent scaffold, including the instructions, available tools,
and how the agent organizes and completes its work. Whereas mini-SWE-agent performs repository
inspection, editing, and testing through its only available tool, Bash, Claude Code and Codex expose
broader tool sets. Claude Code, for example, provides dedicated tools for searching, reading, and
editing files. Both model-specific harnesses used the same SWE-Serve task and verifier setup,
hardware allocation, closed-book policy, and 210-minute wall-time limit. The 350-step limit applied
to mini-SWE-agent was not imposed on the model-specific harnesses.

For both evaluated configurations, GPT-5.6 Sol and Claude Opus 5, the model-specific harness has
lower mean wall-clock time but higher mean inference cost than mini-SWE-agent
(Figure~\ref{fig:code-agent-efficiency}), with lower observed mean pass@1, as reported in
Section~\ref{sec:agent-harness-sensitivity}. For GPT-5.6 Sol,
Codex has lower mean uncached input-token use but higher mean output-token use; Claude Code shows
the opposite pattern for Claude Opus 5 (Figure~\ref{fig:code-agent-tokens}).
Faster execution is not necessarily associated with higher pass@1, lower inference cost, or fewer
output tokens. For the two evaluated configurations, these results support using mini-SWE-agent for
the primary leaderboard and show that measured performance and resource use differ across harnesses.

\clearpage

\subsection{Limitations of executable scoring}
\label{app:scoring-limitations}

\begin{figure}[H]
  \centering
  \includegraphics[width=\linewidth]{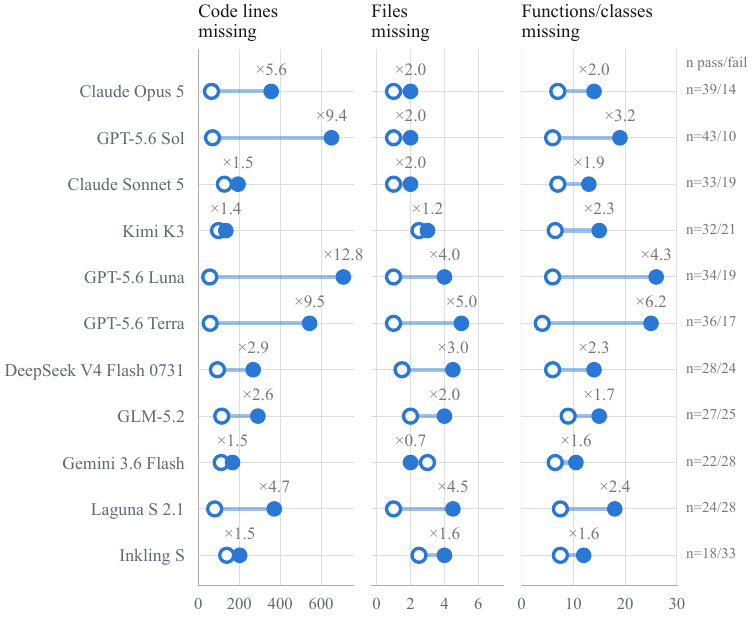}
  \caption{Missing scope differences relative to oracle patches for passing and failing agent-created
  patches from the first SWE-Serve run attempts of the 11 top-per-model configurations.
  Median missing scope is reported relative to the oracle patch in code lines, files, and
  estimated changed functions/classes. Hollow points show
  medians for passing attempts and filled points show medians for failed attempts.
  $\times$ reports the ratio of failed-attempt to passing-attempt medians for each scope measurement.
  $n$ reports the
  number of usable patches (passing/failing) for each of the 11 top-per-model configurations.
  Configurations are ordered by overall mean pass@1 to match Table~\ref{tab:primary-results};
  scope measurements use only the first-run attempts, a subset of the attempts used to calculate
  that mean.}
  \label{fig:appendix-scope-dumbbell}
\end{figure}

As discussed in Section~\ref{sec:limitations}, passing the verifier does not establish every dimension
of upstream readiness. Comparing an agent-created patch with the oracle patch also does not establish
upstream readiness: the oracle is one reference implementation, and alternative correct solutions may
differ in scope without being lower quality. However, because oracle patches derive from merged
upstream changes, they provide a useful reference for examining implementation scope beyond
executable scores. Figure~\ref{fig:appendix-scope-dumbbell} and Table~\ref{tab:appendix-patch-scope}
compare agent-created patches with oracle patch scope as coarse indicators of implementation
complexity. Figure~\ref{fig:appendix-scope-dumbbell} visualizes missing scope, while
Table~\ref{tab:appendix-patch-scope} reports exact median measurements of both missing and extra
scope relative to the oracle patches. These measurements can highlight differences for further
review, but do not assess maintainability or architectural fit, nor replace maintainer review.

Before examining verifier outcomes, we selected for analysis the first run for each of the 11
top-per-model configurations and considered all task attempts from that run.
Figure~\ref{fig:appendix-scope-dumbbell} and Table~\ref{tab:appendix-patch-scope} include all attempts
with both a usable patch capture and a verifier pass/fail result. This excludes nine attempts:
one has an unusable patch capture, and eight encountered agent-created patch or agent errors
that prevented the verifier from producing a pass/fail result. These exclusions apply to
individual attempts rather than removing the corresponding tasks across all configurations.

\clearpage
We align agent-created and oracle patches by file path and measure missing and extra scope
relative to the oracle patch in production code lines, files, and estimated changed functions/classes.
We estimate how many functions and classes each patch changes in two ways. First, we identify
function and class names in diff headers, which can identify the function or class containing a
change even when its declaration is unchanged. Second, we identify added or removed lines that
declare a function or class. Within each file, each distinct function name and each distinct class
name is counted once, even if it appears in multiple changes. Missing and extra scope measure
differences in counts rather than identifying specific missing functionality. We then calculate
median measurements separately for passing and failing attempts within each configuration.
For every configuration, failed attempts have a higher median missing-line count than passing
attempts. However, these groups contain different tasks, so the comparison does not isolate the
relationship between patch scope and verifier outcome on the same tasks.

\begin{table}[H]
  \caption{Patch scope differences relative to oracle patches for passing and failing agent-created
  patches from the first SWE-Serve run attempts of the 11 top-per-model configurations.
  Median missing and extra scope is reported relative to the oracle patch in code lines, files,
  and estimated changed functions/classes; $n$ reports the number of usable patches (passing/failing)
  for each of the 11 top-per-model configurations. Configurations are ordered by overall mean pass@1
  to match Table~\ref{tab:primary-results}; scope measurements use only the first-run attempts,
  a subset of the attempts used to calculate that mean.}
  \label{tab:appendix-patch-scope}
  \vspace{6pt}
  \centering
  \begingroup
  \small
  \setlength{\tabcolsep}{3pt}
  \renewcommand{\arraystretch}{1.08}
  \begin{tabular*}{\linewidth}{@{\extracolsep{\fill}}llc*{6}{>{\centering\arraybackslash}p{3.8em}}@{}}
    \toprule
    Model & Outcome & $n$ & \multicolumn{2}{c}{Code lines}
          & \multicolumn{2}{c}{Files} & \multicolumn{2}{c@{}}{Functions/classes} \\
    \cmidrule(lr){4-5}\cmidrule(lr){6-7}\cmidrule(lr){8-9}
          & & & Missing & Extra & Missing & Extra & Missing & Extra \\
    \midrule
\input{generated/patch_scope_rows.tex} 
    \bottomrule
  \end{tabular*}
  \endgroup
\end{table}

\clearpage

%% file: generated/inference_taxonomy_matrix.tex
% Generated by scripts/paper/generate_inference_taxonomy_table.py; do not edit.
% Source manifest SHA-256: 04eb64ded1182e04ba9e1f1dcbf5bd7955b0cf744e47d897516406af300df216
\begin{table}[!t]
  \setlength{\tabcolsep}{3pt}
  \renewcommand{\arraystretch}{1.15}
  \caption{Evidence for constructing the runtime-domain taxonomy from four prominent open-source inference systems. Documented components illustrate four recurring responsibilities: request handling and I/O, scheduling and request lifecycle, model execution, and KV cache and runtime-resource management. Bracketed keys link to the corresponding evidence in pinned repository versions.}
  \label{tab:inference-taxonomy-components}
  \begin{center}
  \normalsize
  \begin{tabular}{@{}>{\raggedright\arraybackslash}p{0.13\textwidth}
                          >{\raggedright\arraybackslash}p{0.20\textwidth}
                          >{\raggedright\arraybackslash}p{0.20\textwidth}
                          >{\raggedright\arraybackslash}p{0.20\textwidth}
                          >{\raggedright\arraybackslash}p{0.20\textwidth}@{}}
    \toprule
    System & Request / I/O & Scheduling /\newline lifecycle & Model\newline execution & KV / runtime\newline resources \\
    \midrule
    \textbf{vLLM} & API server and input/output processors [\href{https://github.com/vllm-project/vllm/blob/0a5a55136fa0552a19816b211c4a57f1a341d830/docs/design/arch_overview.md?plain=1\#L71-L77}{V-IO1}, \href{https://github.com/vllm-project/vllm/blob/0a5a55136fa0552a19816b211c4a57f1a341d830/docs/design/arch_overview.md?plain=1\#L141-L155}{V-IO2}] & EngineCore scheduler [\href{https://github.com/vllm-project/vllm/blob/0a5a55136fa0552a19816b211c4a57f1a341d830/docs/design/arch_overview.md?plain=1\#L79-L85}{V-SL}] & Executor, GPU worker, ModelRunner [\href{https://github.com/vllm-project/vllm/blob/0a5a55136fa0552a19816b211c4a57f1a341d830/docs/design/arch_overview.md?plain=1\#L87-L93}{V-ME1}, \href{https://github.com/vllm-project/vllm/blob/0a5a55136fa0552a19816b211c4a57f1a341d830/docs/design/arch_overview.md?plain=1\#L181-L185}{V-ME2}] & KV cache manager [\href{https://github.com/vllm-project/vllm/blob/0a5a55136fa0552a19816b211c4a57f1a341d830/vllm/v1/core/kv_cache_manager.py\#L118-L188}{V-RM}] \\
    \addlinespace[4pt]\arrayrulecolor{black!25}\midrule[0.3pt]\arrayrulecolor{black}\addlinespace[4pt]
    \textbf{SGLang} & HTTP server, TokenizerManager, \mbox{Detokenizer}\allowbreak{}\mbox{Manager} [\href{https://github.com/sgl-project/sglang/blob/f825d729363136a2d4a4b330fa694d0b37a878fa/python/sglang/srt/entrypoints/engine.py\#L207-L218}{S-IO}] & Scheduler, request queues, batch construction [\href{https://github.com/sgl-project/sglang/blob/f825d729363136a2d4a4b330fa694d0b37a878fa/python/sglang/srt/entrypoints/engine.py\#L211-L214}{S-SL1}, \href{https://github.com/sgl-project/sglang/blob/f825d729363136a2d4a4b330fa694d0b37a878fa/python/sglang/srt/managers/scheduler.py\#L1144-L1150}{S-SL2}] & TP worker, ModelRunner, attention and kernels [\href{https://github.com/sgl-project/sglang/blob/f825d729363136a2d4a4b330fa694d0b37a878fa/python/sglang/srt/managers/tp_worker.py\#L463-L477}{S-ME1}, \href{https://github.com/sgl-project/sglang/blob/f825d729363136a2d4a4b330fa694d0b37a878fa/python/sglang/srt/model_executor/model_runner.py\#L284-L350}{S-ME2}, \href{https://github.com/sgl-project/sglang/blob/f825d729363136a2d4a4b330fa694d0b37a878fa/python/sglang/srt/model_executor/model_runner.py\#L919-L939}{S-ME3}, \href{https://github.com/sgl-project/sglang/blob/f825d729363136a2d4a4b330fa694d0b37a878fa/python/sglang/srt/model_executor/model_runner.py\#L1219-L1220}{S-ME4}] & KV pools, allocators, RadixCache [\href{https://github.com/sgl-project/sglang/blob/f825d729363136a2d4a4b330fa694d0b37a878fa/python/sglang/srt/mem_cache/memory_pool.py\#L15-L20}{S-RM1}, \href{https://github.com/sgl-project/sglang/blob/f825d729363136a2d4a4b330fa694d0b37a878fa/python/sglang/srt/mem_cache/radix_cache.py\#L303-L329}{S-RM2}] \\
    \addlinespace[4pt]\arrayrulecolor{black!25}\midrule[0.3pt]\arrayrulecolor{black}\addlinespace[4pt]
    \textbf{TensorRT-\newline LLM} & LLM interface and \mbox{tokenization/}\allowbreak{}\mbox{detokenization} [\href{https://github.com/NVIDIA/TensorRT-LLM/blob/e189237f0c3d3070af2e33488c3f3ab579a4f231/docs/source/torch/arch_overview.md?plain=1\#L6-L15}{T-IO}] & PyExecutor scheduler [\href{https://github.com/NVIDIA/TensorRT-LLM/blob/e189237f0c3d3070af2e33488c3f3ab579a4f231/docs/source/torch/arch_overview.md?plain=1\#L17-L31}{T-SL}] & Model engine and decoder [\href{https://github.com/NVIDIA/TensorRT-LLM/blob/e189237f0c3d3070af2e33488c3f3ab579a4f231/docs/source/torch/arch_overview.md?plain=1\#L33-L41}{T-ME}] & ResourceManager and KVCacheManager [\href{https://github.com/NVIDIA/TensorRT-LLM/blob/e189237f0c3d3070af2e33488c3f3ab579a4f231/docs/source/torch/arch_overview.md?plain=1\#L54-L69}{T-RM}] \\
    \addlinespace[4pt]\arrayrulecolor{black!25}\midrule[0.3pt]\arrayrulecolor{black}\addlinespace[4pt]
    \textbf{Triton} & HTTP/REST, gRPC, C API, returning outputs [\href{https://github.com/triton-inference-server/server/blob/e1db8a603d863b1317ee859bdeea2be19959462e/docs/user_guide/architecture.md?plain=1\#L31-L54}{R-IO}] & Per-model schedulers and batchers [\href{https://github.com/triton-inference-server/server/blob/e1db8a603d863b1317ee859bdeea2be19959462e/docs/user_guide/architecture.md?plain=1\#L34-L44}{R-SL}] & Model backends and instances [\href{https://github.com/triton-inference-server/server/blob/e1db8a603d863b1317ee859bdeea2be19959462e/docs/user_guide/architecture.md?plain=1\#L39-L50}{R-ME1}, \href{https://github.com/triton-inference-server/server/blob/e1db8a603d863b1317ee859bdeea2be19959462e/docs/user_guide/model_execution.md?plain=1\#L53-L64}{R-ME2}] & Sequence state and backend-managed resources [\href{https://github.com/triton-inference-server/server/blob/e1db8a603d863b1317ee859bdeea2be19959462e/docs/user_guide/implicit_state_management.md?plain=1\#L29-L69}{R-RM1}, \href{https://github.com/triton-inference-server/server/blob/e1db8a603d863b1317ee859bdeea2be19959462e/docs/user_guide/implicit_state_management.md?plain=1\#L230-L252}{R-RM2}] \\
    \bottomrule
  \end{tabular}
  \end{center}
\end{table}

%% file: generated/task_characterization_rows.tex
% Generated by scripts/paper/generate_dataset_tables.py; do not edit.
    Adaptive EAGLE step control +\allowbreak{} config guards & Speculative / decoding & CPU & -- & Single\newline \mbox{(M)} & \featurecheck & -- & -- \\
    \arrayrulecolor{black!20}\midrule[0.25pt]\arrayrulecolor{black}
    NVFP4 and DeepSeek MoE inference optimizations & Kernels / quant. / perf. & CPU & -- & Multi\newline \mbox{(M+K)} & \featurecheck & -- & -- \\
    \arrayrulecolor{black!20}\midrule[0.25pt]\arrayrulecolor{black}
    Gemma 4 fused Triton operations & Kernels / quant. / perf. & GPU & -- & Single\newline \mbox{(M)} & -- & -- & -- \\
    \arrayrulecolor{black!20}\midrule[0.25pt]\arrayrulecolor{black}
    Masked MoE activation +\allowbreak{} GDN QKV-split kernels & Kernels / quant. / perf. & GPU & -- & Single\newline \mbox{(M)} & -- & -- & -- \\
    \arrayrulecolor{black!20}\midrule[0.25pt]\arrayrulecolor{black}
    HiCache eviction and tombstone correctness & Caching / runtime state & GPU & -- & Multi\newline \mbox{(S+K)} & \featurecheck & -- & -- \\
    \arrayrulecolor{black!20}\midrule[0.25pt]\arrayrulecolor{black}
    HiCache framework +\allowbreak{} sliding-window attention & Caching / runtime state & GPU & -- & Multi\newline \mbox{(S+K)} & \featurecheck & -- & -- \\
    \arrayrulecolor{black!20}\midrule[0.25pt]\arrayrulecolor{black}
    Custom spec-algorithm registry +\allowbreak{} N-gram metric fix & Speculative / decoding & GPU & -- & Single\newline \mbox{(M)} & \featurecheck & -- & -- \\
    \arrayrulecolor{black!20}\midrule[0.25pt]\arrayrulecolor{black}
    Anthropic Messages request/response conversion & Serving API / runtime & CPU & -- & Single\newline \mbox{(I)} & -- & -- & -- \\
    \arrayrulecolor{black!20}\midrule[0.25pt]\arrayrulecolor{black}
    Spec: forward-timeout ordering fix (EAGLE v1) & Speculative / decoding & GPU & \featurecheck & Multi\newline \mbox{(S+M)} & \featurecheck & -- & -- \\
    \arrayrulecolor{black!20}\midrule[0.25pt]\arrayrulecolor{black}
    Mamba state offloading +\allowbreak{} HybridCacheController & Caching / runtime state & GPU & -- & Single\newline \mbox{(K)} & \featurecheck & -- & -- \\
    \arrayrulecolor{black!20}\midrule[0.25pt]\arrayrulecolor{black}
    DeepSeek-V4 DSML tool-call parser & Serving API / runtime & GPU & -- & Single\newline \mbox{(I)} & \featurecheck & -- & -- \\
    \arrayrulecolor{black!20}\midrule[0.25pt]\arrayrulecolor{black}
    O(1) EAGLE bigram RadixKey view & Speculative / decoding & CPU & -- & Single\newline \mbox{(K)} & \featurecheck & -- & -- \\
    \arrayrulecolor{black!20}\midrule[0.25pt]\arrayrulecolor{black}
    torch.mm for DeepSeek-V3.2 indexer GEMM & Kernels / quant. / perf. & GPU & -- & Single\newline \mbox{(M)} & -- & -- & \featurecheck \\
    \arrayrulecolor{black!20}\midrule[0.25pt]\arrayrulecolor{black}
    Spec: split accept\_length counters & Speculative / decoding & CPU & -- & Single\newline \mbox{(M)} & \featurecheck & -- & -- \\
    \arrayrulecolor{black!20}\midrule[0.25pt]\arrayrulecolor{black}
    Kimi-K2.5 EAGLE3-MLA draft path & Speculative / decoding & CPU & -- & Single\newline \mbox{(M)} & -- & -- & -- \\
    \arrayrulecolor{black!20}\midrule[0.25pt]\arrayrulecolor{black}
    Spec: split draft-extend dataclass & Speculative / decoding & CPU & -- & Single\newline \mbox{(M)} & \featurecheck & -- & -- \\
    \arrayrulecolor{black!20}\midrule[0.25pt]\arrayrulecolor{black}
    PD hybrid state-transfer refactor & Distributed / scheduling & CPU & -- & Single\newline \mbox{(K)} & -- & -- & -- \\
    \arrayrulecolor{black!20}\midrule[0.25pt]\arrayrulecolor{black}
    DeepSeek-V4 w4a4 MegaMoE & Model / backend & GPU & -- & Single\newline \mbox{(M)} & \featurecheck & -- & -- \\
    \arrayrulecolor{black!20}\midrule[0.25pt]\arrayrulecolor{black}
    PD disaggregation priority-scheduling fix & Distributed / scheduling & GPU & -- & Single\newline \mbox{(S)} & \featurecheck & -- & -- \\
    \arrayrulecolor{black!20}\midrule[0.25pt]\arrayrulecolor{black}
    Kimi tokenizer TTFT fast path & Kernels / quant. / perf. & GPU & -- & Single\newline \mbox{(I)} & -- & -- & \featurecheck \\
    \arrayrulecolor{black!20}\midrule[0.25pt]\arrayrulecolor{black}
    UnifiedRadixCache device-match semantics fix & Caching / runtime state & GPU & -- & Single\newline \mbox{(K)} & \featurecheck & -- & -- \\
    \arrayrulecolor{black!20}\midrule[0.25pt]\arrayrulecolor{black}
    TMA bulk-store MLA KV-buffer kernel & Kernels / quant. / perf. & GPU & -- & Multi\newline \mbox{(M+K)} & \featurecheck & -- & \featurecheck \\
    \arrayrulecolor{black!20}\midrule[0.25pt]\arrayrulecolor{black}
    Spec-V2 paged tree drafting & Speculative / decoding & GPU & -- & Single\newline \mbox{(K)} & -- & -- & -- \\
    \arrayrulecolor{black!20}\midrule[0.25pt]\arrayrulecolor{black}
    Ideogram4 NVFP4 config from safetensors & Model / backend & CPU & -- & Single\newline \mbox{(M)} & -- & -- & -- \\
    \arrayrulecolor{black!20}\midrule[0.25pt]\arrayrulecolor{black}
    DFLASH v1 non-overlap serving & Speculative / decoding & GPU & \featurecheck & Multi\newline \mbox{(S+M+K)} & \featurecheck & -- & -- \\
    \arrayrulecolor{black!20}\midrule[0.25pt]\arrayrulecolor{black}
    EAGLE v2 page-one tree runtime & Speculative / decoding & GPU & \featurecheck & Multi\newline \mbox{(S+M+K)} & \featurecheck & \featurecheck & -- \\
    \arrayrulecolor{black!20}\midrule[0.25pt]\arrayrulecolor{black}
    Diffusion-LLM radix-graph serving & Model / backend & GPU & \featurecheck & Multi\newline \mbox{(S+M+K)} & \featurecheck & \featurecheck & -- \\
    \arrayrulecolor{black!20}\midrule[0.25pt]\arrayrulecolor{black}
    DeepSeek-V3.2 index-cache store +\allowbreak{} paged gather & Caching / runtime state & GPU & -- & Multi\newline \mbox{(M+K)} & \featurecheck & -- & -- \\
    \arrayrulecolor{black!20}\midrule[0.25pt]\arrayrulecolor{black}
    SDAR dense +\allowbreak{} MoE day-0 serving & Model / backend & GPU & \featurecheck & Single\newline \mbox{(M)} & -- & -- & -- \\
    \arrayrulecolor{black!20}\midrule[0.25pt]\arrayrulecolor{black}
    EPD registration cleanup & Distributed / scheduling & CPU & -- & Multi\newline \mbox{(I+S)} & \featurecheck & -- & -- \\
    \arrayrulecolor{black!20}\midrule[0.25pt]\arrayrulecolor{black}
    Context-parallel strategy abstractions & Distributed / scheduling & CPU & -- & Single\newline \mbox{(M)} & \featurecheck & -- & -- \\
    \arrayrulecolor{black!20}\midrule[0.25pt]\arrayrulecolor{black}
    Chunked SGMV CUDA-graph replay fix & Serving API / runtime & GPU & -- & Multi\newline \mbox{(M+K)} & \featurecheck & -- & -- \\
    \arrayrulecolor{black!20}\midrule[0.25pt]\arrayrulecolor{black}
    Qwen3.5 dense +\allowbreak{} MoE core serving & Model / backend & GPU & \featurecheck & Multi\newline \mbox{(I+M)} & -- & -- & -- \\
    \arrayrulecolor{black!20}\midrule[0.25pt]\arrayrulecolor{black}
    Breakable prefill CUDA-graph engine & Kernels / quant. / perf. & GPU & -- & Multi\newline \mbox{(M+K)} & \featurecheck & -- & -- \\
    \arrayrulecolor{black!20}\midrule[0.25pt]\arrayrulecolor{black}
    HiSparse page-aware decode lifecycle +\allowbreak{} FP8 routing & Caching / runtime state & GPU & -- & Multi\newline \mbox{(S+M+K)} & \featurecheck & -- & -- \\
    \arrayrulecolor{black!20}\midrule[0.25pt]\arrayrulecolor{black}
    MoE LoRA route alignment +\allowbreak{} numerical kernel & Kernels / quant. / perf. & GPU & -- & Single\newline \mbox{(M)} & -- & -- & -- \\
    \arrayrulecolor{black!20}\midrule[0.25pt]\arrayrulecolor{black}
    Mixed-chunk piecewise CUDA-graph replay & Serving API / runtime & GPU & \featurecheck & Multi\newline \mbox{(S+M+K)} & \featurecheck & \featurecheck & -- \\
    \arrayrulecolor{black!20}\midrule[0.25pt]\arrayrulecolor{black}
    Adaptive speculative allocation safety & Speculative / decoding & CPU & -- & Single\newline \mbox{(K)} & \featurecheck & -- & -- \\
    \arrayrulecolor{black!20}\midrule[0.25pt]\arrayrulecolor{black}
    Transformers-backend compute runtime & Model / backend & GPU & \featurecheck & Single\newline \mbox{(M)} & -- & -- & -- \\
    \arrayrulecolor{black!20}\midrule[0.25pt]\arrayrulecolor{black}
    Transformers 5 MoE core serving & Model / backend & GPU & \featurecheck & Single\newline \mbox{(M)} & \featurecheck & -- & -- \\
    \arrayrulecolor{black!20}\midrule[0.25pt]\arrayrulecolor{black}
    Gemma 4 piecewise CUDA-graph execution & Model / backend & GPU & \featurecheck & Multi\newline \mbox{(I+M+K)} & \featurecheck & -- & -- \\
    \arrayrulecolor{black!20}\midrule[0.25pt]\arrayrulecolor{black}
    Gemma 4 speculative execution & Model / backend & GPU & \featurecheck & Multi\newline \mbox{(S+M+K)} & \featurecheck & -- & -- \\
    \arrayrulecolor{black!20}\midrule[0.25pt]\arrayrulecolor{black}
    Gemma 4 26B-A4B core serving & Model / backend & GPU & \featurecheck & Multi\newline \mbox{(I+M)} & -- & -- & -- \\
    \arrayrulecolor{black!20}\midrule[0.25pt]\arrayrulecolor{black}
    Multi-item Score API packing & Serving API / runtime & GPU & -- & Multi\newline \mbox{(I+S+M)} & -- & \featurecheck & -- \\
    \arrayrulecolor{black!20}\midrule[0.25pt]\arrayrulecolor{black}
    N-gram trie refactor +\allowbreak{} stateful matching & Speculative / decoding & GPU & \featurecheck & Multi\newline \mbox{(S+M+K)} & \featurecheck & -- & -- \\
    \arrayrulecolor{black!20}\midrule[0.25pt]\arrayrulecolor{black}
    UnifiedRadixCache streaming-session KV & Caching / runtime state & GPU & -- & Multi\newline \mbox{(S+K)} & \featurecheck & -- & -- \\
    \arrayrulecolor{black!20}\midrule[0.25pt]\arrayrulecolor{black}
    GLM-V repetition-penalty sampling & Serving API / runtime & GPU & \featurecheck & Single\newline \mbox{(M)} & \featurecheck & -- & -- \\
    \arrayrulecolor{black!20}\midrule[0.25pt]\arrayrulecolor{black}
    N-gram external SAM corpus & Speculative / decoding & GPU & \featurecheck & Multi\newline \mbox{(I+M+K)} & \featurecheck & -- & -- \\
    \arrayrulecolor{black!20}\midrule[0.25pt]\arrayrulecolor{black}
    Embedding-model LoRA serving & Model / backend & GPU & -- & Multi\newline \mbox{(I+M)} & -- & -- & -- \\
    \arrayrulecolor{black!20}\midrule[0.25pt]\arrayrulecolor{black}
    Score API hybrid token +\allowbreak{} embedding inputs & Serving API / runtime & GPU & \featurecheck & Multi\newline \mbox{(I+M)} & -- & -- & -- \\
    \arrayrulecolor{black!20}\midrule[0.25pt]\arrayrulecolor{black}
    Sequence\allowbreak{}Classification Score API & Model / backend & GPU & \featurecheck & Multi\newline \mbox{(I+S+M)} & -- & -- & -- \\
    \arrayrulecolor{black!20}\midrule[0.25pt]\arrayrulecolor{black}
    Spec-V2 streaming-session KV commit & Speculative / decoding & GPU & \featurecheck & Multi\newline \mbox{(S+M+K)} & \featurecheck & \featurecheck & -- \\
    \arrayrulecolor{black!20}\midrule[0.25pt]\arrayrulecolor{black}
    EAGLE3 overlap metadata lifecycle & Serving API / runtime & GPU & \featurecheck & Multi\newline \mbox{(S+M)} & \featurecheck & \featurecheck & -- \\

%% file: generated/task_validation_coverage_rows.tex
% Generated by scripts/paper/generate_reviewer_task_validation_ledger.py; do not edit.
No-op rejection & 53/53 \\
Oracle acceptance & 53/53 \\
Test coverage and scoring review (F2P/P2P) & 53/53 \\
Task-specific agent-assisted adversarial review of candidate solutions & 53/53 \\
Additional causal review for tasks triggering a zero-pass screen & 9/9 (all applicable) \\

%% file: generated/openbook_retrieval_by_model_rows.tex
    \multicolumn{5}{@{}l}{\textbf{Any external fetch}} \\
    \texttt{xhigh} & 41.5\% & 37.7\% & 34.0\% & 54.7\% \\
    \texttt{high} & 30.2\% & 11.3\% & 13.2\% & 34.0\% \\
    \texttt{medium} & 15.1\% & 13.2\% &  3.8\% & 17.0\% \\
    \addlinespace[0.8em]
    \multicolumn{5}{@{}l}{\textbf{Confirmed out-of-band retrieval}} \\
    \texttt{xhigh} & 13.2\% & 17.0\% & 15.1\% & 32.1\% \\
    \texttt{high} & 15.1\% &  5.7\% &  5.7\% & 15.1\% \\
    \texttt{medium} &  3.8\% &  3.8\% &  0.0\% &  3.8\% \\

%% file: generated/consistency_result_rows.tex
% Generated by scripts/paper/generate_consistency_artifacts.py; do not edit.
    GPT-5.6 Sol & \texttt{max} & 75.5\% & 83.0\% & 67.9\% \\
    Claude Opus 5 & \texttt{max} & 75.5\% & 81.1\% & 69.8\% \\
    Claude Opus 5 & \texttt{high} & 73.6\% & 79.2\% & 66.0\% \\
    Claude Opus 5 & \texttt{xhigh} & 72.3\% & 81.1\% & 62.3\% \\
    Claude Opus 5 & \texttt{medium} & 68.6\% & 77.4\% & 58.5\% \\
    GPT-5.6 Sol & \texttt{xhigh} & 66.0\% & 75.5\% & 52.8\% \\
    GPT-5.6 Terra & \texttt{max} & 64.2\% & 77.4\% & 54.7\% \\
    Claude Sonnet 5 & \texttt{xhigh} & 64.2\% & 75.5\% & 50.9\% \\
    Kimi K3 & \texttt{max} & 64.2\% & 75.5\% & 50.9\% \\
    GPT-5.6 Luna & \texttt{max} & 64.2\% & 73.6\% & 50.9\% \\
    Claude Sonnet 5 & \texttt{max} & 61.0\% & 71.7\% & 50.9\% \\
    Claude Opus 5 & \texttt{low} & 60.4\% & 67.9\% & 54.7\% \\
    Claude Sonnet 5 & \texttt{high} & 57.9\% & 69.8\% & 43.4\% \\
    GPT-5.6 Sol & \texttt{high} & 56.0\% & 69.8\% & 43.4\% \\
    DeepSeek V4 Flash (0731) & \texttt{max} & 55.3\% & 73.6\% & 41.5\% \\
    Claude Sonnet 5 & \texttt{medium} & 55.3\% & 69.8\% & 39.6\% \\
    GPT-5.6 Luna & \texttt{xhigh} & 55.3\% & 64.2\% & 45.3\% \\
    GPT-5.6 Sol & \texttt{medium} & 54.1\% & 66.0\% & 37.7\% \\
    GPT-5.6 Terra & \texttt{xhigh} & 52.8\% & 64.2\% & 43.4\% \\
    GPT-5.6 Terra & \texttt{high} & 49.1\% & 62.3\% & 34.0\% \\
    Claude Sonnet 5 & \texttt{low} & 49.1\% & 60.4\% & 35.8\% \\
    GLM-5.2 & \texttt{max} & 48.4\% & 64.2\% & 34.0\% \\
    Gemini 3.6 Flash & \texttt{high} & 47.8\% & 60.4\% & 30.2\% \\
    Laguna S 2.1 & \texttt{max} & 45.9\% & 64.2\% & 26.4\% \\
    GPT-5.6 Luna & \texttt{high} & 45.3\% & 56.6\% & 35.8\% \\
    GPT-5.6 Terra & \texttt{medium} & 44.7\% & 54.7\% & 34.0\% \\
    GPT-5.6 Terra & \texttt{low} & 44.0\% & 54.7\% & 35.8\% \\
    GPT-5.6 Sol & \texttt{low} & 40.9\% & 52.8\% & 28.3\% \\
    GPT-5.6 Luna & \texttt{medium} & 37.7\% & 49.1\% & 24.5\% \\
    Inkling S & \texttt{xhigh} & 34.6\% & 45.3\% & 24.5\% \\
    GPT-5.6 Luna & \texttt{low} & 20.8\% & 28.3\% & 11.3\% \\

%% file: generated/strict_e2e_task_decomposition_rows.tex
\textbf{\shortstack[l]{Tasks with non-E2E\\tests (15)}} & & & & & & \\
\addlinespace[2pt]
Spec: forward-timeout ordering fix (EAGLE v1) & 1 & 13 & 31 & 0 & 1 & 1 \\
\arrayrulecolor{black!20}\specialrule{0.25pt}{0.3pt}{0.3pt}\arrayrulecolor{black}
DFLASH v1 non-overlap serving & 34 & 3 & 0 & 0 & 32 & 1 \\
\arrayrulecolor{black!20}\specialrule{0.25pt}{0.3pt}{0.3pt}\arrayrulecolor{black}
EAGLE v2 page-one tree runtime & 19 & 2 & 6 & 25 & 2 & 0 \\
\arrayrulecolor{black!20}\specialrule{0.25pt}{0.3pt}{0.3pt}\arrayrulecolor{black}
Diffusion-LLM radix-graph serving & 8 & 8 & 0 & 2 & 31 & 0 \\
\arrayrulecolor{black!20}\specialrule{0.25pt}{0.3pt}{0.3pt}\arrayrulecolor{black}
SDAR dense + MoE day-0 serving & 2 & 3 & 10 & 17 & 6 & 0 \\
\arrayrulecolor{black!20}\specialrule{0.25pt}{0.3pt}{0.3pt}\arrayrulecolor{black}
Qwen3.5 dense + MoE core serving & 16 & 5 & 9 & 16 & 8 & 0 \\
\arrayrulecolor{black!20}\specialrule{0.25pt}{0.3pt}{0.3pt}\arrayrulecolor{black}
Transformers-backend compute runtime & 3 & 2,277 & 4 & 0 & 28 & 1 \\
\arrayrulecolor{black!20}\specialrule{0.25pt}{0.3pt}{0.3pt}\arrayrulecolor{black}
Transformers 5 MoE core serving & 4 & 1 & 21 & 7 & 0 & 5 \\
\arrayrulecolor{black!20}\specialrule{0.25pt}{0.3pt}{0.3pt}\arrayrulecolor{black}
Gemma 4 speculative execution & 4 & 1 & 26 & 7 & 0 & 0 \\
\arrayrulecolor{black!20}\specialrule{0.25pt}{0.3pt}{0.3pt}\arrayrulecolor{black}
Gemma 4 26B-A4B core serving & 3 & 2 & 16 & 16 & 1 & 0 \\
\arrayrulecolor{black!20}\specialrule{0.25pt}{0.3pt}{0.3pt}\arrayrulecolor{black}
N-gram trie refactor + stateful matching & 4 & 34 & 28 & 0 & 3 & 2 \\
\arrayrulecolor{black!20}\specialrule{0.25pt}{0.3pt}{0.3pt}\arrayrulecolor{black}
N-gram external SAM corpus & 2 & 22 & 8 & 5 & 20 & 0 \\
\arrayrulecolor{black!20}\specialrule{0.25pt}{0.3pt}{0.3pt}\arrayrulecolor{black}
SequenceClassification Score API & 3 & 37 & 14 & 0 & 16 & 3 \\
\arrayrulecolor{black!20}\specialrule{0.25pt}{0.3pt}{0.3pt}\arrayrulecolor{black}
Spec-V2 streaming-session KV commit & 32 & 15 & 3 & 0 & 30 & 0 \\
\arrayrulecolor{black!20}\specialrule{0.25pt}{0.3pt}{0.3pt}\arrayrulecolor{black}
EAGLE3 overlap metadata lifecycle & 110 & 7 & 15 & 18 & 0 & 0 \\
\midrule
\textbf{\shortstack[l]{Tasks without non-E2E\\tests (4)}} & & & & & & \\
\addlinespace[2pt]
Mixed-chunk piecewise CUDA-graph replay & 3 & 0 & 15 & 18 & 0 & 0 \\
\arrayrulecolor{black!20}\specialrule{0.25pt}{0.3pt}{0.3pt}\arrayrulecolor{black}
Gemma 4 piecewise CUDA-graph execution & 6 & 0 & 29 & 4 & 0 & 0 \\
\arrayrulecolor{black!20}\specialrule{0.25pt}{0.3pt}{0.3pt}\arrayrulecolor{black}
GLM-V repetition-penalty sampling & 13 & 0 & 27 & 6 & 0 & 0 \\
\arrayrulecolor{black!20}\specialrule{0.25pt}{0.3pt}{0.3pt}\arrayrulecolor{black}
Score API hybrid token + embedding inputs & 9 & 0 & 26 & 6 & 0 & 1 \\
\midrule
Total & 276 & 2,430 & 288 & 147 & 178 & 14 \\

%% file: generated/patch_scope_rows.tex
% Generated by scripts/paper/generate_model_result_tables.py; do not edit.
    Claude Opus 5 & Pass & 39 & $64$ & $82$ & $1$ & $0$ & $7$ & $7$ \\
     & Fail & 14 & $356$ & $156.5$ & $2$ & $2$ & $14$ & $15.5$ \\
    \arrayrulecolor{black!20}\specialrule{0.25pt}{2pt}{2pt}\arrayrulecolor{black}
    GPT-5.6 Sol & Pass & 43 & $69$ & $54$ & $1$ & $1$ & $6$ & $2$ \\
     & Fail & 10 & $651$ & $58.5$ & $2$ & $0.5$ & $19$ & $5$ \\
    \arrayrulecolor{black!20}\specialrule{0.25pt}{2pt}{2pt}\arrayrulecolor{black}
    Claude Sonnet 5 & Pass & 33 & $127$ & $31$ & $1$ & $0$ & $7$ & $1$ \\
     & Fail & 19 & $194$ & $28$ & $2$ & $0$ & $13$ & $1$ \\
    \arrayrulecolor{black!20}\specialrule{0.25pt}{2pt}{2pt}\arrayrulecolor{black}
    Kimi K3 & Pass & 32 & $98.5$ & $39$ & $2.5$ & $0$ & $6.5$ & $2$ \\
     & Fail & 21 & $134$ & $114$ & $3$ & $1$ & $15$ & $4$ \\
    \arrayrulecolor{black!20}\specialrule{0.25pt}{2pt}{2pt}\arrayrulecolor{black}
    GPT-5.6 Luna & Pass & 34 & $55.5$ & $53.5$ & $1$ & $0.5$ & $6$ & $2.5$ \\
     & Fail & 19 & $709$ & $196$ & $4$ & $1$ & $26$ & $8$ \\
    \arrayrulecolor{black!20}\specialrule{0.25pt}{2pt}{2pt}\arrayrulecolor{black}
    GPT-5.6 Terra & Pass & 36 & $57.5$ & $75$ & $1$ & $1$ & $4$ & $3.5$ \\
     & Fail & 17 & $544$ & $175$ & $5$ & $1$ & $25$ & $9$ \\
    \arrayrulecolor{black!20}\specialrule{0.25pt}{2pt}{2pt}\arrayrulecolor{black}
    DeepSeek V4 Flash (0731) & Pass & 28 & $93$ & $32.5$ & $1.5$ & $0$ & $6$ & $1$ \\
     & Fail & 24 & $268$ & $37.5$ & $4.5$ & $0$ & $14$ & $2$ \\
    \arrayrulecolor{black!20}\specialrule{0.25pt}{2pt}{2pt}\arrayrulecolor{black}
    GLM-5.2 & Pass & 27 & $114$ & $21$ & $2$ & $0$ & $9$ & $0$ \\
     & Fail & 25 & $291$ & $21$ & $4$ & $0$ & $15$ & $0$ \\
    \arrayrulecolor{black!20}\specialrule{0.25pt}{2pt}{2pt}\arrayrulecolor{black}
    Gemini 3.6 Flash & Pass & 22 & $111.5$ & $26$ & $3$ & $0$ & $6.5$ & $0$ \\
     & Fail & 28 & $167$ & $43$ & $2$ & $0.5$ & $10.5$ & $1$ \\
    \arrayrulecolor{black!20}\specialrule{0.25pt}{2pt}{2pt}\arrayrulecolor{black}
    Laguna S 2.1 & Pass & 24 & $79.5$ & $29$ & $1$ & $0$ & $7.5$ & $1$ \\
     & Fail & 28 & $371$ & $43.5$ & $4.5$ & $1$ & $18$ & $3$ \\
    \arrayrulecolor{black!20}\specialrule{0.25pt}{2pt}{2pt}\arrayrulecolor{black}
    Inkling S & Pass & 18 & $138.5$ & $4.5$ & $2.5$ & $0$ & $7.5$ & $0$ \\
     & Fail & 33 & $202$ & $4$ & $4$ & $0$ & $12$ & $1$ \\